\documentclass[runningheads]{llncs}
\usepackage[T1]{fontenc}
\usepackage{graphicx}
\usepackage{svg}
\usepackage{subfig}
\usepackage{caption}
\usepackage{amsmath}
\usepackage{booktabs}
\usepackage{multirow}
\usepackage{tabularx}
\usepackage{array}
\usepackage{amssymb}
\usepackage{transparent}
\usepackage{makecell}
\usepackage{xurl}
\usepackage{hyperref}
\usepackage{siunitx}
\usepackage{graphicx}
\usepackage{adjustbox}

\begin{document}
\title{WallZero: Mastering the Game of WallGo with Strategic Analysis}
\author{
Hsing-Yu Chen\inst{1,2}\orcidID{0009-0007-1471-5225} \and
Jérôme Arjonilla\inst{2}\orcidID{0000-0002-0082-1939} \and
I-Chen Wu\inst{1}\orcidID{0000-0003-2535-0587} \and
Ti-Rong Wu\inst{2}\orcidID{0000-0002-7532-3176}}
\authorrunning{Chen et al.}
\institute{National Yang Ming Chiao Tung University, Hsinchu, Taiwan \and
Academia Sinica, Taipei, Taiwan\\
\email{hsingyu.cs14@nycu.edu.tw, jerome@iis.sinica.edu.tw, \\ icwu@nycu.edu.tw, tirongwu@iis.sinica.edu.tw}
}

\maketitle              

\setcounter{footnote}{0}

\begin{abstract}
WallGo is a recently introduced strategic board game popularized by the 2025 Netflix series \emph{The Devil’s Plan}. 
Although played on a small $7 \times 7$ board, its combination of stone movement and wall placement yields high game-tree complexity and intricate strategic interactions.
Despite its growing popularity, WallGo remains underexplored.
This paper presents \textit{WallZero}, an AlphaZero-based agent for the two-player WallGo setting. 
We introduce tailored action and feature designs to improve playing performance significantly.
In the evaluation, WallZero defeats two professional Go players who participated in this study, securing on average 1.98$\times$ more territory per game.
Beyond its strength, we use WallZero to assess game fairness and identify key strategies for mastering WallGo. 
Interestingly, our results show that the opening used in the Netflix series yields a more balanced game.
Our code is available at https://rlg.iis.sinica.edu.tw/papers/wallzero.

\end{abstract}

\section{Introduction}

\textit{WallGo} is one of the strategic board games introduced in the 2025 Netflix series \textit{The Devil’s Plan}.
Following its release in May 2025, the series ranked within Netflix’s Global Top 10 for non-English programs and accumulated over 1.7 million views in its first week~\cite{_top_}.
Notably, the legendary Go player Lee Sedol--known for his historic matches against AlphaGo--also participated in the show, further drawing attention from both the Go and AI communities.

Among the various games presented in the series, WallGo attracted substantial discussion in online forums, and many enthusiasts attempted to implement the game and develop AI agents for it~\cite{chu_schaoss_2025,team_play_}.
The game can be played by two to four players on a 7×7 board, where the objective is to enclose more territory than the opponents. 
Although the board size is small, each turn combines stone movement and wall construction, resulting in an estimated game-tree complexity of approximately $10^{87}$.\footnote{The estimate is based on the two-player version, assuming an average branching factor of 64 and an average game depth of 48 measured from 2,000 self-play games generated by last 50 snapshot models of WallZero.}
Despite this growing interest, WallGo has not been thoroughly explored from an AI perspective.
Most existing online agents have not reached a high level of play, and it remains unclear whether AI can achieve superhuman performance in WallGo.
Moreover, the game's strategies and balance have not been systematically analyzed.
Inspired by prior works applying AlphaZero to analyze board game properties~\cite{tomasev_assessing_2020,wang_evaluating_2025}, we develop an AlphaZero-based agent, named \textit{WallZero}, to study WallGo.
WallZero incorporates tailored action and feature designs to enhance playing strength. 
Empirical evaluation shows that WallZero outperforms two professional Go players who participated in this study.
Furthermore, we use WallZero to analyze the balance of different game modes and to uncover key strategies for mastering WallGo. 
Together, our results demonstrate the potential of AlphaZero-based methods for mastering and systematically understanding newly introduced board games.

\section{Background}

\subsection{Rules of WallGo}

\textit{WallGo} is a competitive board game for two to four players, played on a $7 \times 7$ grid.
Similar to the game of Go, the objective is to enclose more territory than the opponent to win. 
This paper focuses on the two-player setting of WallGo, where \textit{Red} (first player) and \textit{Blue} (second player) each control four identical stones.
The game proceeds in two phases: the \textit{setup phase} and the \textit{play phase}.
During the setup phase, each player places their remaining stones onto the board.
The game then transitions to the play phase, in which players alternately move their stones and construct walls to partition the board and enclose territory.
The two phases are described in detail below.

\begin{figure}[h]
    \centering
    \captionsetup[subfigure]{justification=centering}
    
    \subfloat[]{
        \includegraphics[width=0.18\columnwidth]{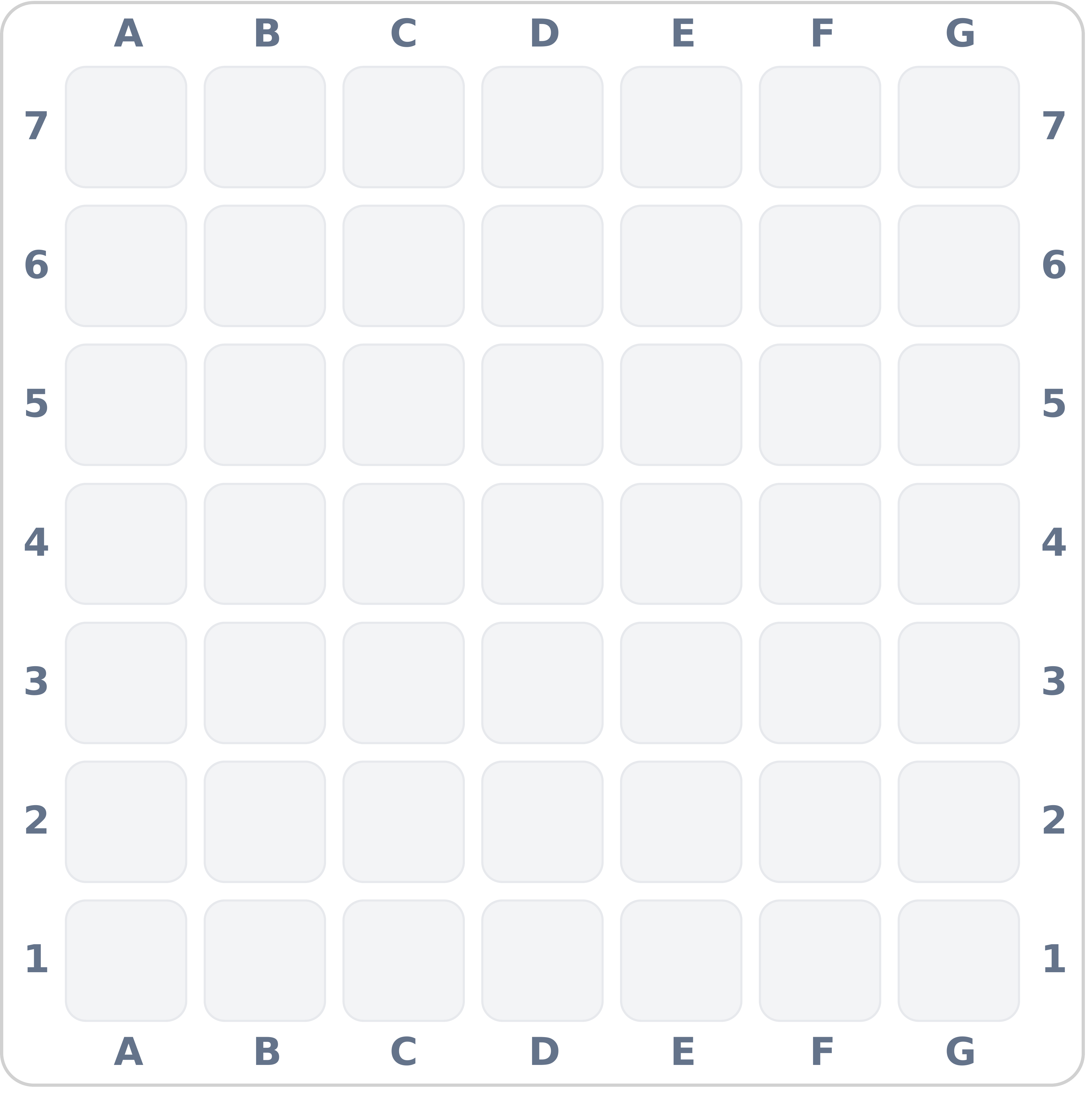}\label{fig:rule-empty}
    }
    \subfloat[]{
        \includegraphics[width=0.18\columnwidth]{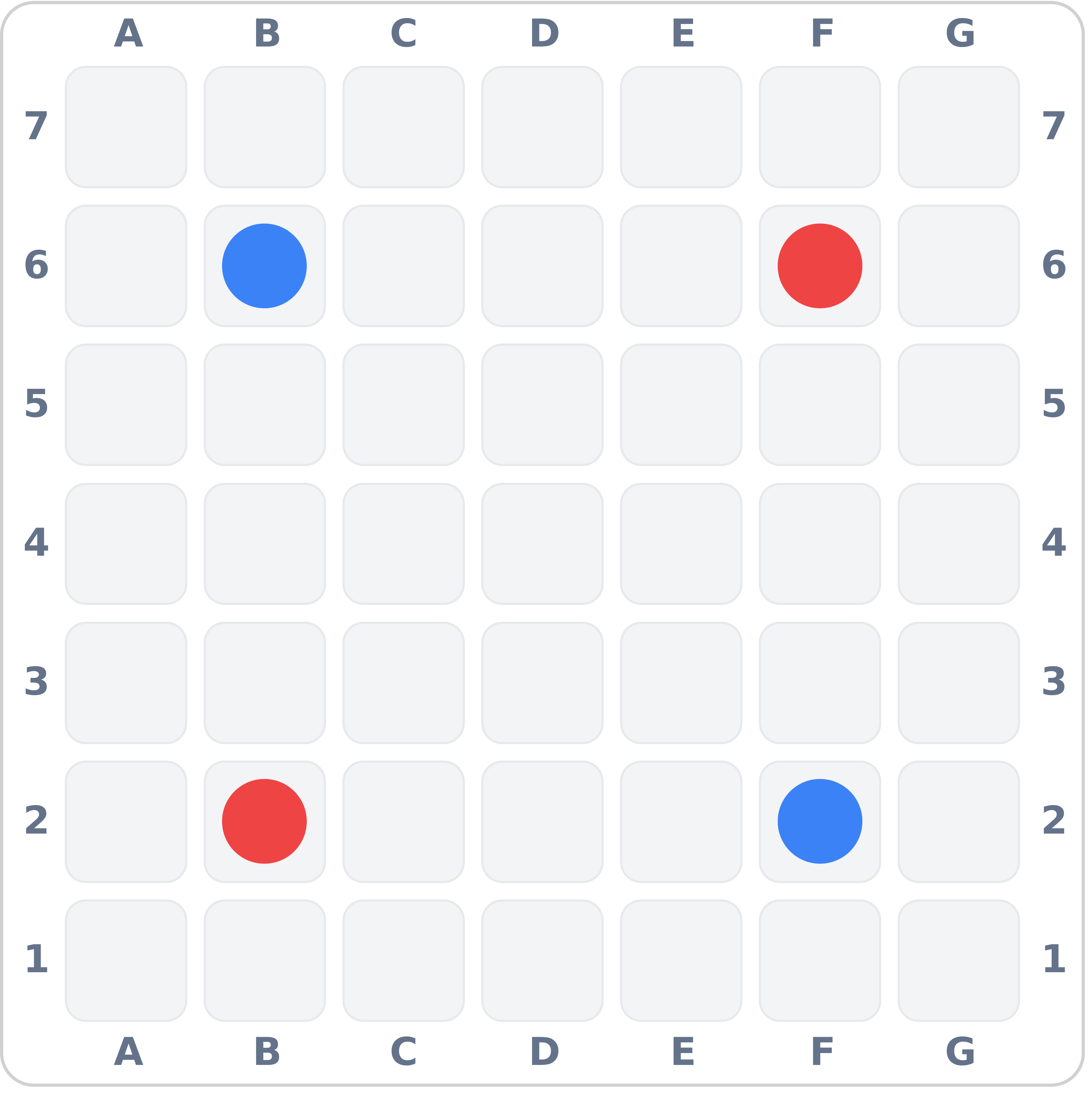}\label{fig:rule-4-stone}
    }
    \subfloat[]{
        \includegraphics[width=0.18\columnwidth]{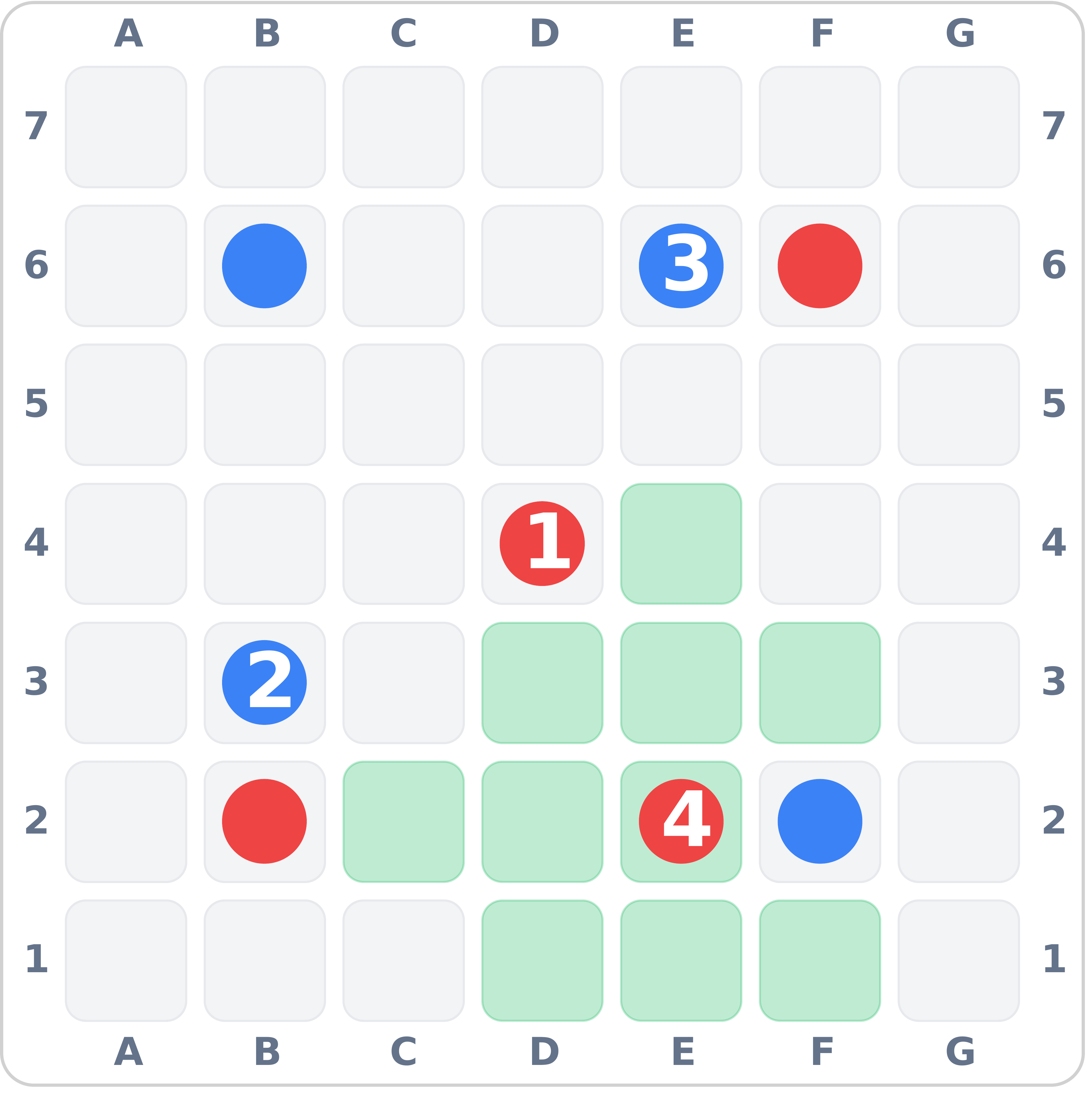}\label{fig:rule-setup}
    }
    \subfloat[]{
        \includegraphics[width=0.18\columnwidth]{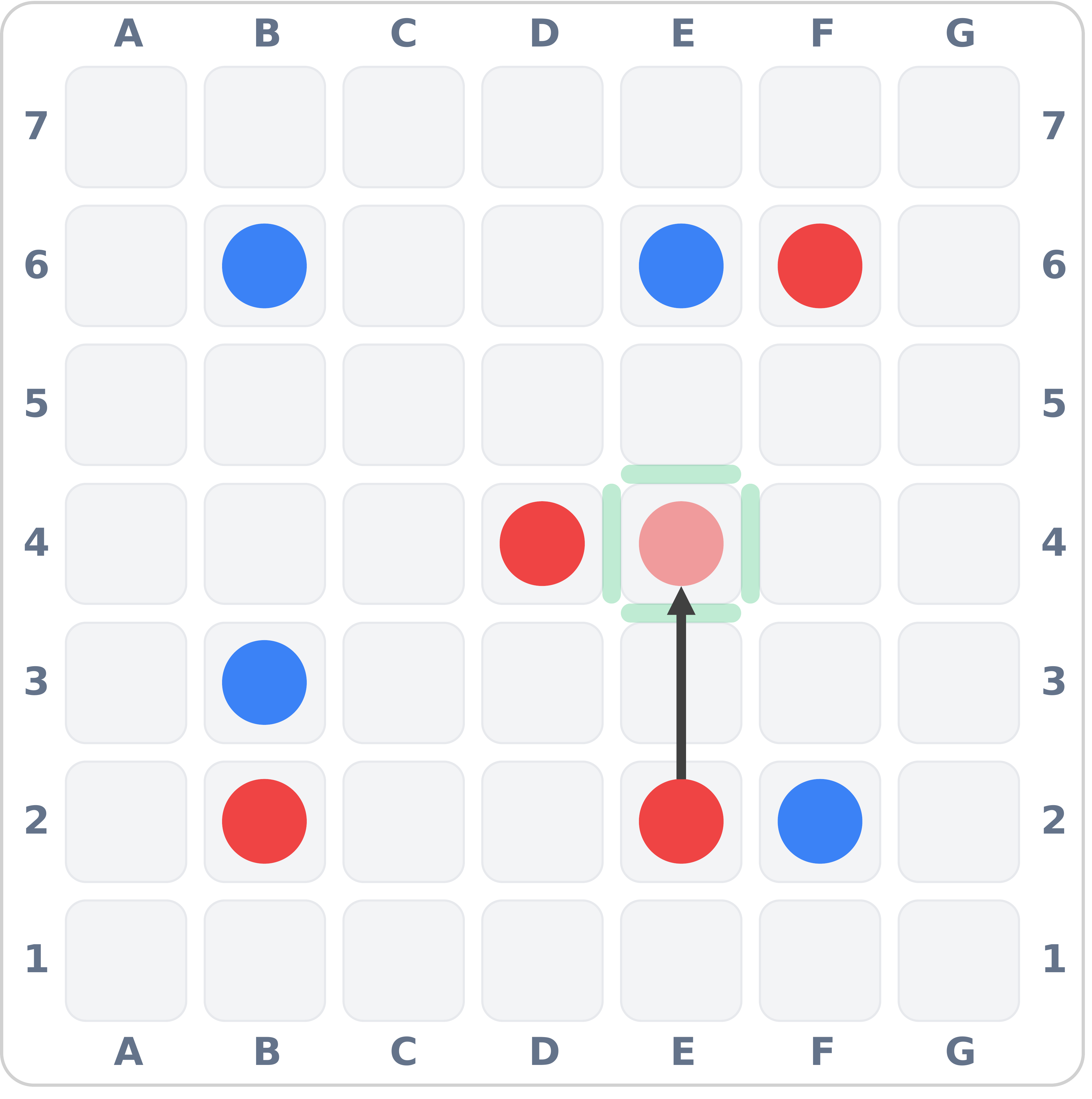}\label{fig:rule-play}
    }
    \subfloat[]{
        \includegraphics[width=0.18\columnwidth]{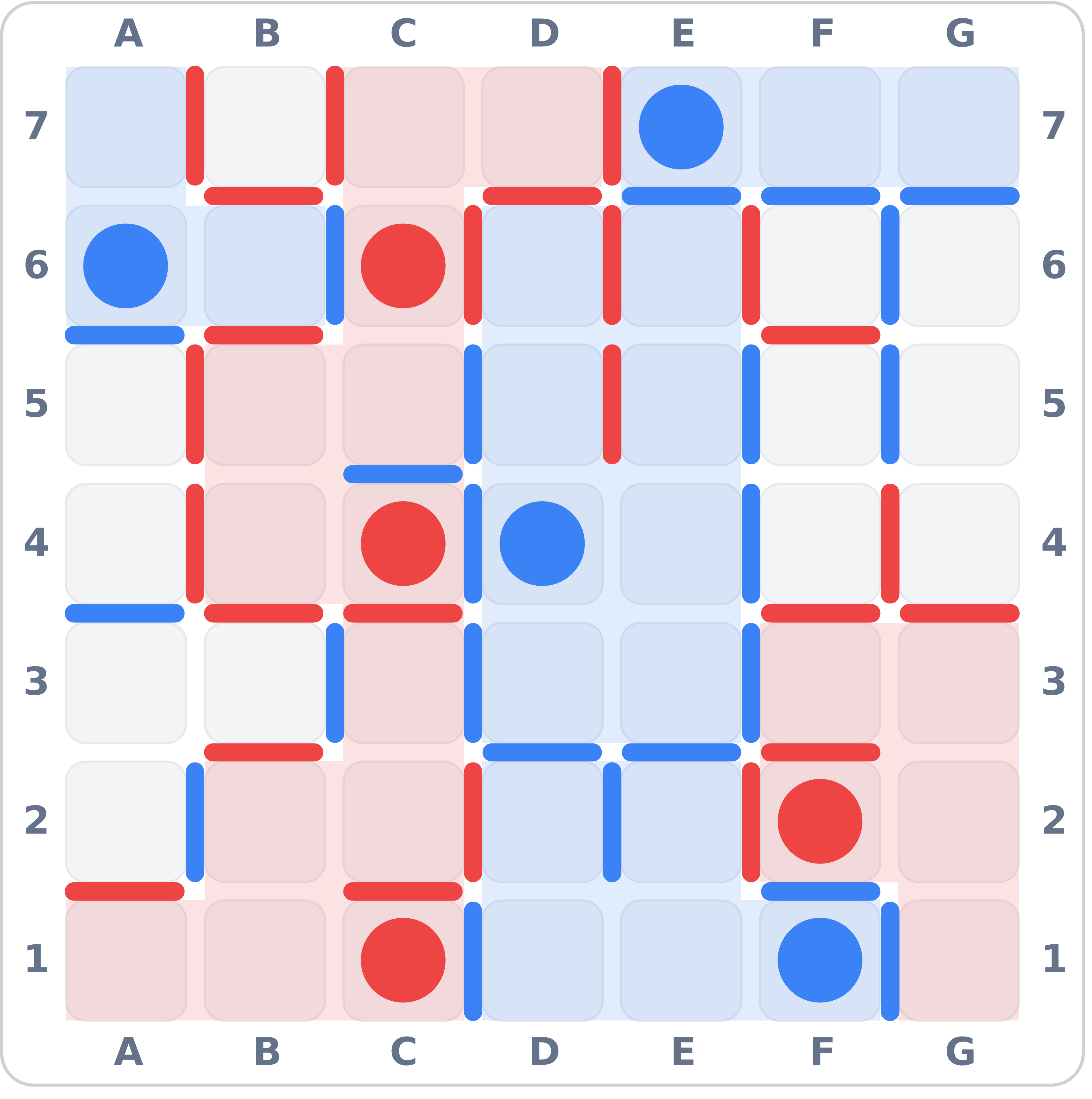}\label{fig:rule-endgame}
    }
    
    \caption{Overview of WallGo rules. (a) Empty mode. (b) 4-stone mode. (c) Reachable area of Red stone labeled 4 (light green). (d) Two-step upward move with wall placement. (e) Endgame: Blue wins by 1 point (19--18).}
    \label{fig:rule}
\end{figure}

\textbf{Setup Phase.}
We consider two initial modes, shown in Figures~\ref{fig:rule-empty} and~\ref{fig:rule-4-stone}.
In the \textit{empty mode}, the board starts empty, and all eight stones are placed in this phase.
In the \textit{4-stone mode}, used in \textit{The Devil's Plan}, four stones are pre-positioned, and each player places their remaining two stones.
In both modes, placement follows a fixed order: Red places one stone first, after which players alternately place two consecutive stones, starting with Blue, until each player has four stones on the board.
Stones are placed on any empty square.
After all stones are placed, the game proceeds to the play phase.

\textbf{Play Phase.}
Players alternate turns, starting with Red.
On each turn, a player must first select one of their stones and move it by zero (i.e., remain in place), one, or two orthogonal steps, and then place one wall.
Movement is step-by-step and cannot pass through walls or other stones.
For example, the Red stone labeled 4 in Figure~\ref{fig:rule-setup} cannot move to the right.
After moving, the player must place one wall along one of the four edges adjacent to the stone's final position, as shown in Figure~\ref{fig:rule-play}.

\textbf{Game End and Scoring.}
An enclosed region belongs to a player if it contains only that player's stones (e.g., C4 and C6 in Figure~\ref{fig:rule-endgame}), and squares within it count toward that player's territory.
Regions containing no stones (e.g., A4, A5) are neutral and count toward neither player.
The game ends when all stones are enclosed within regions containing stones of only one player.
The player with the larger total territory wins.
If totals are equal, the player with the largest single region wins; otherwise, the game is a draw.

\subsection{AlphaZero}

AlphaZero~\cite{silver_mastering_2017,silver_general_2018} is a general-purpose reinforcement learning algorithm that integrates deep neural networks with Monte Carlo Tree Search (MCTS)~\cite{kocsis_bandit_2006,coulom_efficient_2007}, enabling agents to achieve superhuman performance across several board games from scratch through self-play without domain-specific knowledge.
In AlphaZero, a residual neural network takes the state $s$ as input and jointly outputs a policy distribution $\pi_\theta(a \mid s)$, which provides prior probabilities over legal actions, and a value $v_\theta(s)$, which estimates the win rate.
During self-play, MCTS uses the policy and value networks to guide the tree search efficiently.

Since its introduction, subsequent studies have highlighted that the design of state representations is crucial for both learning efficiency and final playing performance.
For example, incorporating game-specific features accelerated training in Go~\cite{wu_accelerating_2020}, while strategically derived features improved playing performance in Chess~\cite{czech_representation_2024}.
Furthermore, AlphaZero-based methods have increasingly been used to assess game balance~\cite{tomasev_assessing_2020} and game difficulty~\cite{wang_evaluating_2025}, as their superhuman strength enables reliable large-scale analysis.

\section{WallZero}
This section presents the design of \textit{WallZero}, an AlphaZero-based agent for playing WallGo.
The training pipeline mostly follows the AlphaZero algorithm, with two WallGo-specific designs: the action design and the feature design.

\subsection{Action Design}

In WallGo, actions differ between phases.
To maintain a unified policy network across phases, we define the action vector space as
\begin{equation*}
|\mathcal{A}| =
\underbrace{49}_{\text{setup phase}}
+
\underbrace{49 \times 13 \times 4}_{\text{play phase}}
= 2597.
\end{equation*}

In the setup phase, players place a stone on a $7 \times 7$ board position. 
In the play phase, a move consists of selecting one of its stones (49 possible positions), choosing a destination (up to 13 reachable positions within two steps), and selecting one of four wall directions. 
Although the action space is fully enumerated, only a subset of actions is legal at each state.
We also apply an action mask to filter out invalid actions from the policy output.

\subsection{Feature Design}
\label{approach:state}

Inspired by previous work demonstrating that feature design plays a crucial role in the performance of AlphaZero-based agents, we design several features tailored to WallGo.
The overall feature design is summarized in Table~\ref{tab:feature_definitions}.
Each feature plane is a binary $7 \times 7$ tensor aligned with the board representation.
We consider four feature configurations: a \textit{Base configuration} (\texttt{B}) and three additional configurations, including \textit{Territory} (\texttt{T}), \textit{Reachability} (\texttt{R}), and \textit{History} (\texttt{H}).

\begin{table}[ht]
\centering
\caption{Feature design in WallZero. The base configuration (\texttt{B}) includes stone, wall, and player turn planes; \texttt{T}, \texttt{R}, and \texttt{H} are extended features.}
\scriptsize
\setlength{\tabcolsep}{6pt}
\begin{adjustbox}{width=\columnwidth}
\label{tab:feature_definitions}
\begin{tabular}{lrl}
\toprule
\textbf{Feature} & \textbf{\# of planes} & \textbf{Description} \\
\midrule

Stone 
& \makecell[r]{2} 
& \makecell[l]{Red / Blue stone} \\

Horizontal Wall 
& \makecell[r]{2}
& \makecell[l]{Red / Blue horizontal wall} \\

Vertical Wall
& \makecell[r]{2}
& \makecell[l]{Red / Blue vertical wall} \\

Player Turn
& \makecell[r]{2}
& \makecell[l]{Indicates the player to move} \\

\midrule

Territory (\texttt{T})
& \makecell[r]{3}
& \makecell[l]{Red / Blue / Neutral territory} \\

Reachability (\texttt{R})
& \makecell[r]{2}
& \makecell[l]{Positions reachable within one turn for each player} \\

History (\texttt{H})
& \makecell[r]{36}
& \makecell[l]{Four-step history of stones, walls, and territory (9 per step)} \\

\bottomrule
\end{tabular}
\end{adjustbox}
\end{table}

\textbf{Base configuration (\texttt{B}):}
The base configuration encodes the necessary information to represent the current board state.
Two planes represent stone positions for Red and Blue, four planes represent horizontal and vertical walls for each player (two per player), and two planes indicate the player to move.
We separate wall features by player to identify the builder of each wall, which provides more information about player-specific strategic intentions.
Since wall locations are fewer than board grids, the remaining entries are padded with zeros to maintain consistent tensor dimensions. 
In total, the base configuration consists of eight feature planes.

\textbf{Territory (\texttt{T}):}
Since territory determines the final score in WallGo, we introduce the Territory feature by adding three planes that encode Red-controlled, Blue-controlled, and neutral regions.
This allows the model to explicitly distinguish different territory ownership.

\textbf{Reachability (\texttt{R}):}
WallGo allows a stone to move zero to two steps per turn, and both stones and walls affect which positions are reachable. 
To capture this property, we add one plane per player representing the union of positions reachable by at least one of that player's stones. 

\textbf{History (\texttt{H}):}
Following AlphaZero, we stack previous frames to provide temporal information about recent board changes.
To limit the total number of feature planes, we consider only a four-step history, where each step includes stone, wall, and territory planes.

In total, the full feature representation used in WallZero contains 49 planes.

\section{Experiments}

\subsection{Setup}

We trained WallZero for both empty and 4-stone modes using the MiniZero framework~\cite{wu_minizero_2025}.
The model consists of residual blocks with 256 hidden channels. 
During the self-play phase, MCTS used 200 simulations per move.
The training spanned 500 iterations, each consisting of 2,000 self-play games and 500 optimization steps. 
Optimization used a batch size of 1,024 and a constant learning rate of 0.02.
All training was conducted on four NVIDIA GTX 1080Ti GPUs with dual Intel E5-2683 v3 CPUs. 

\subsection{Feature Design Performance}

We evaluate the feature designs described in Section~\ref{approach:state}, including WallZero-\texttt{B}, WallZero-\texttt{BT}, WallZero-\texttt{BTR}, and WallZero-\texttt{BTRH} (hereafter referred to as WallZero), where the suffix denotes the included features.
In addition, we apply data augmentation via board symmetries in WallZero, including rotations and reflections.
Each model is a 1-block residual network trained for approximately 456 GPU-hours.
After training, we conducted a round-robin tournament in which each pair played 1,000 games (500 as Red and 500 as Blue).

\begin{table}[ht]
\centering
\caption{
Win rates with 95\% confidence intervals across models in empty and 4-stone modes.
Avg. denotes the average win rate against the other three models in the round-robin tournament.}
\label{tab:win_rate_empty_mode}
\scriptsize

\setlength{\tabcolsep}{6pt}
\begin{adjustbox}{width=\columnwidth}
\begin{tabular}{l|cccc}
\toprule
 & \textbf{WallZero-\texttt{B}} & \textbf{WallZero-\texttt{BT}} & \textbf{WallZero-\texttt{BTR}} & \textbf{WallZero} \\ 
\midrule

\textbf{Empty Mode (Avg.)} 
& \makecell[c]{10.65$\pm$1.10\%} 
& \makecell[c]{27.05$\pm$1.59\%} 
& \makecell[c]{79.43$\pm$1.45\%} 
& \makecell[c]{82.87$\pm$1.35\%} \\

\midrule

\textbf{4-Stone Mode (Avg.)} 
& \makecell[c]{10.45$\pm$1.09\%} 
& \makecell[c]{28.65$\pm$1.62\%} 
& \makecell[c]{78.88$\pm$1.46\%} 
& \makecell[c]{82.02$\pm$1.37\%} \\

\bottomrule
\end{tabular}
\end{adjustbox}
\end{table}

The results, in Table~\ref{tab:win_rate_empty_mode}, demonstrate that the feature design strongly influences the agent's performance under the same training time.
Both modes show a similar trend: the inclusion of the reachability feature leads to the most substantial improvement. 
For example, in empty mode, the win rate increases from 27.05\% (WallZero-\texttt{BT}) to 79.43\% (WallZero-\texttt{BTR}).
In addition, the model using full features (WallZero) achieves the best performance (82.87\% in empty mode and 82.02\% in 4-stone mode), and this feature design is used in all subsequent experiments in this paper. 

Next, we analyzed whether either player has an advantage. 
Across 5,000 WallZero self-play games for each mode, Red achieves a win rate of 53.57\% $\pm$ 1.38\% in empty mode, indicating a slight advantage.
In contrast, the win rate for Red is 50.37\% ± 1.38\% in 4-stone mode, suggesting that the trained WallZero agents exhibit a more balanced outcome under the rule adopted in \textit{The Devil's Plan}.

\subsection{Human-AI Evaluation}
\label{subsec:exp_human_ai}

To further evaluate WallZero's strength, we invited two Taiwanese professional Go players--Wei Huang (3-dan) and Chun-Hsun Chou (9-dan)--to compete against WallZero. 
For this, we extended the model to a 10-block residual network and adopted a two-stage curriculum: pre-training on data generated by the 1-block model with the same optimization steps as the 1-block model, followed by 500 iterations. 
This 10-block model required approximately 49 GPU-hours for pre-training and 902 GPU-hours for subsequent training. 

Before the formal matches, both players were given the opportunity to play several practice games until they were familiar with the rules and game strategy.
In the formal matches, each player competed as both Red and Blue in both game modes, resulting in four matches per player. 
Following the rules used in \textit{The Devil's Plan}, a 90-second time limit per move was applied. 
For all matches, WallZero uses 2,000 simulations per move.

\begin{table}[ht]
\centering
\caption{Game results between professional Go players and WallZero. Scores are reported as Human--WallZero territory counts.
Values in parentheses denote the ratio of WallZero's territory to the human's.}

\label{tab:AlphaZero-performance}
\scriptsize

\setlength{\tabcolsep}{9.8pt}
\begin{adjustbox}{width=\columnwidth}
\begin{tabular}{cccccc}
\toprule

& \multicolumn{2}{c}{Empty Mode} && \multicolumn{2}{c}{4-Stone Mode} \\

\noalign{\vspace{1pt}} \cline{2-3} \cline{5-6} \noalign{\vspace{1pt}}
& Red & Blue && Red & Blue \\
\midrule
3-dan & 16--33 (2.06$\times$) & 20--29 (1.45$\times$) && 14--32 (2.29$\times$) & 19--30 (1.58$\times$) \\
9-dan & 12--37 (3.08$\times$) & 17--30 (1.76$\times$) && 20--29 (1.45$\times$) & 12--34 ($2.83\times$) \\

\bottomrule
\end{tabular}
\end{adjustbox}
\end{table}

Table~\ref{tab:AlphaZero-performance} shows that WallZero won all eight matches. 
Remarkably, it secured on average (geometric mean) 1.98$\times$ more territory than the professional Go players per game, and in the most extreme case, it achieved up to 3.08$\times$ against the 9-dan player.
Additionally, its estimated win rate exceeded $90\%$ before the 20th move for all games, regardless of whether it played Red or Blue.
These results indicate that WallZero achieves a level exceeding that of the evaluated professional Go players.
Post-match feedback from professional Go players further noted that wall construction is particularly challenging, since each wall simultaneously restricts both players and requires careful long-term reachability planning.

\subsection{Analysis of Opening Configurations}

This section analyzes openings in empty and 4-stone modes. 
All evaluations use the 10-block residual models introduced in Section~\ref{subsec:exp_human_ai}.

\subsubsection{Empty Mode to 4-Stone Opening Analysis.} 

\begin{figure}[ht]
    \centering
    \fontsize{6}{7}\selectfont
    \captionsetup[subfigure]{justification=centering}

    \subfloat[Step 0\\55.95\%]{
        \includegraphics[width=0.15\textwidth]{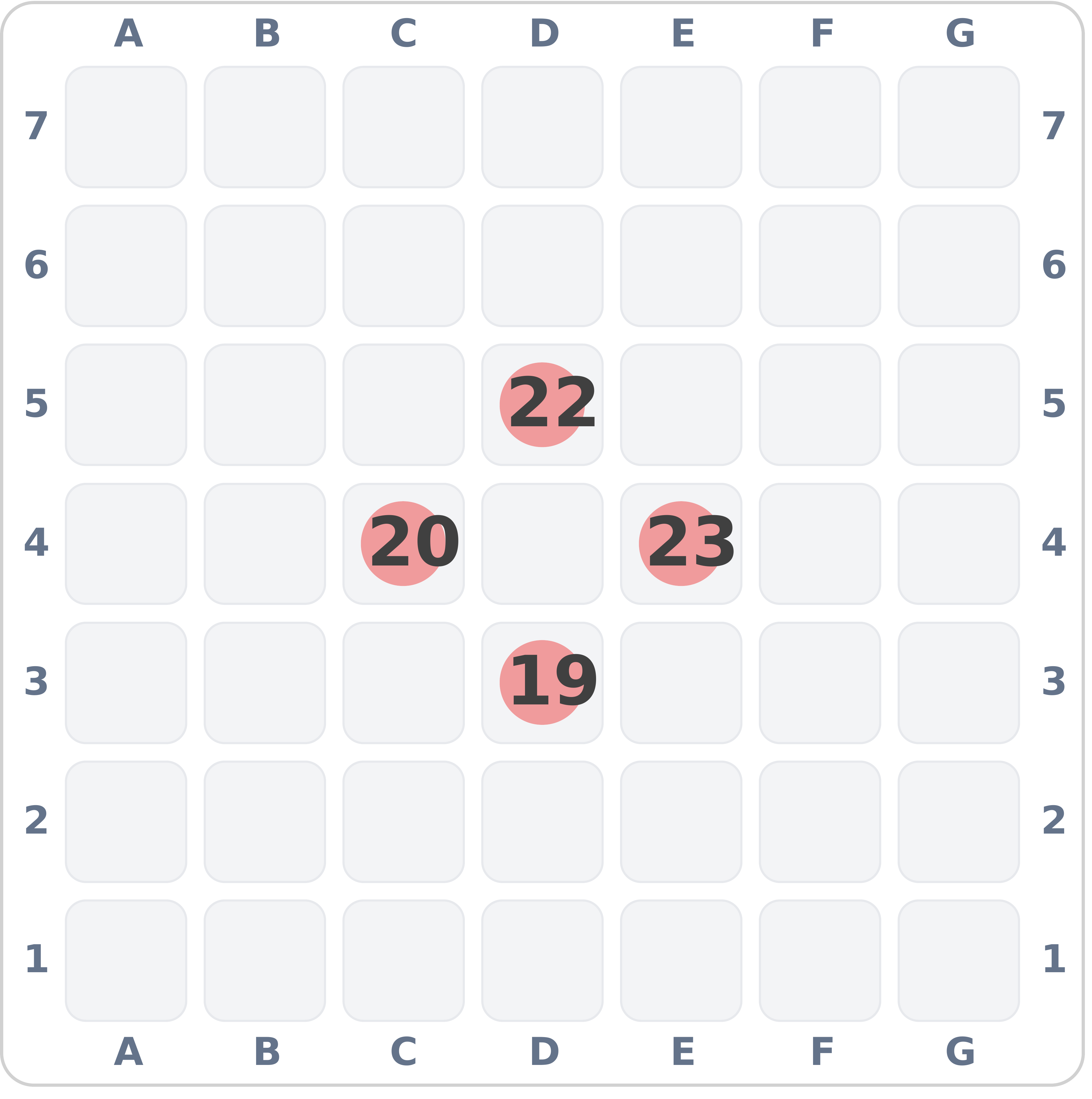}\label{fig:empty_step1}
    }
    \subfloat[Step 1\\49.6\%]{
        \includegraphics[width=0.15\textwidth]{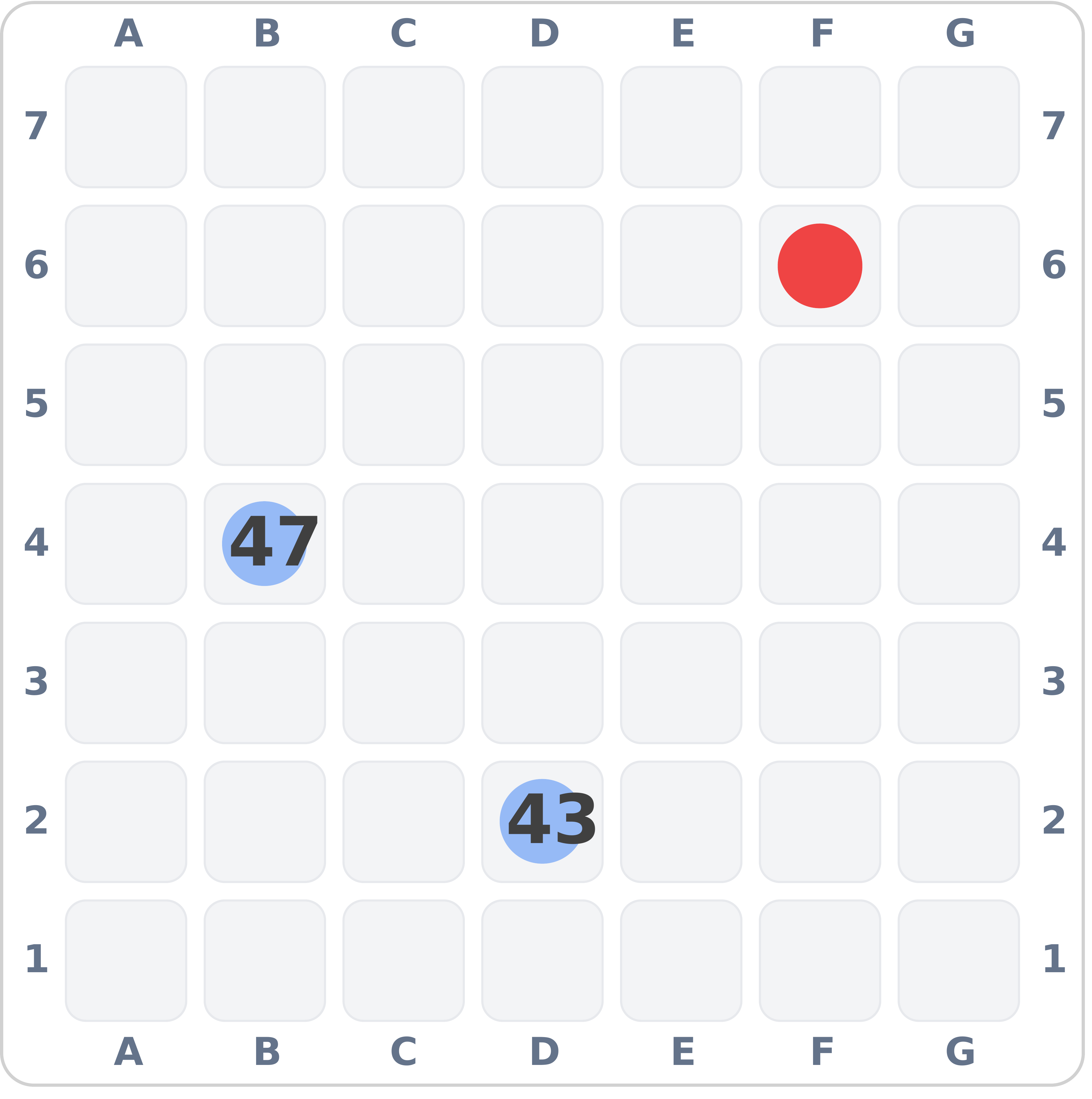}\label{fig:empty_step2}
    }
    \subfloat[Step 2\\54.7\%]{
        \includegraphics[width=0.15\textwidth]{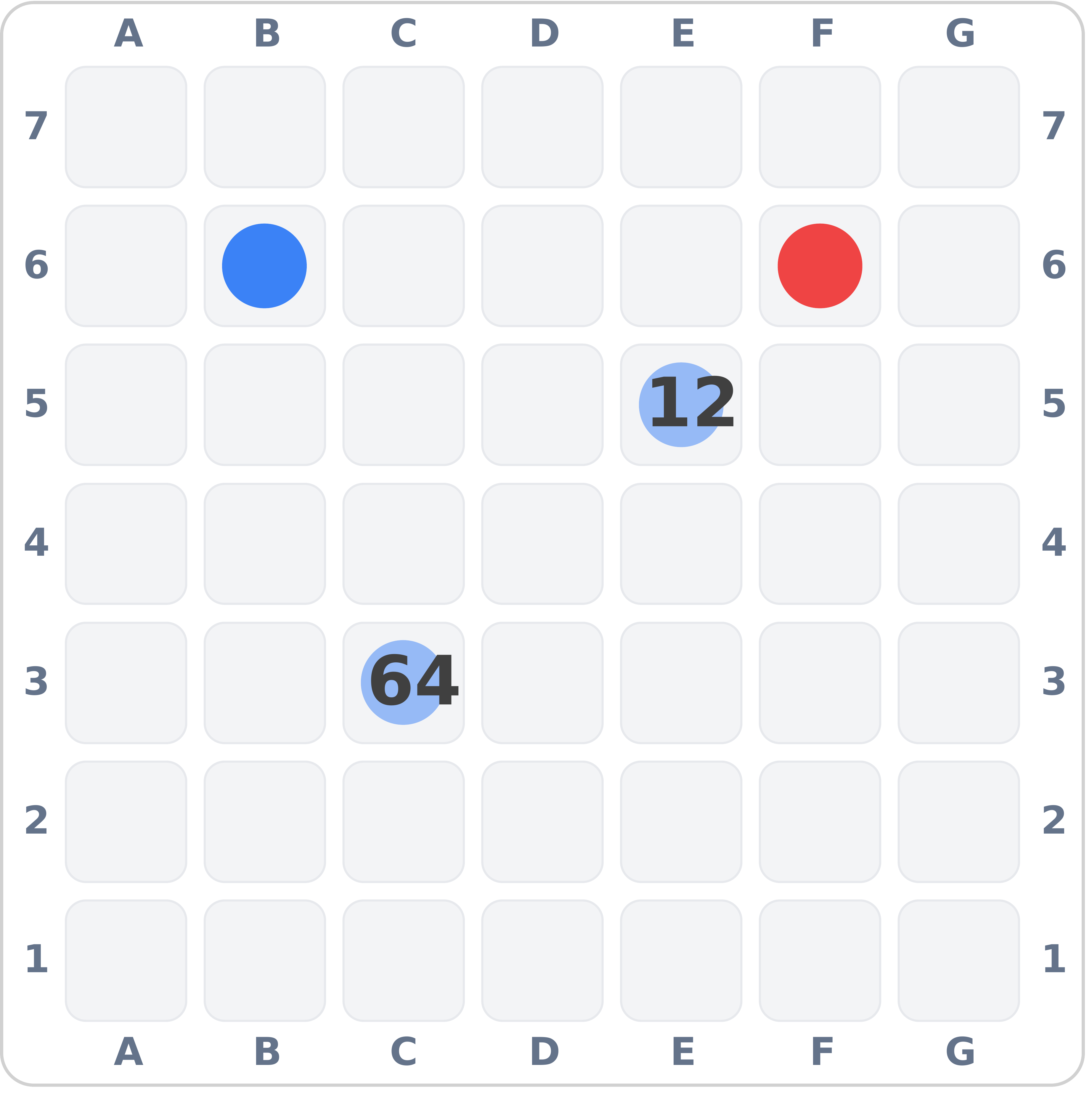}\label{fig:empty_step3}
    }
    \subfloat[Step 3\\52.95\%]{
        \includegraphics[width=0.15\textwidth]{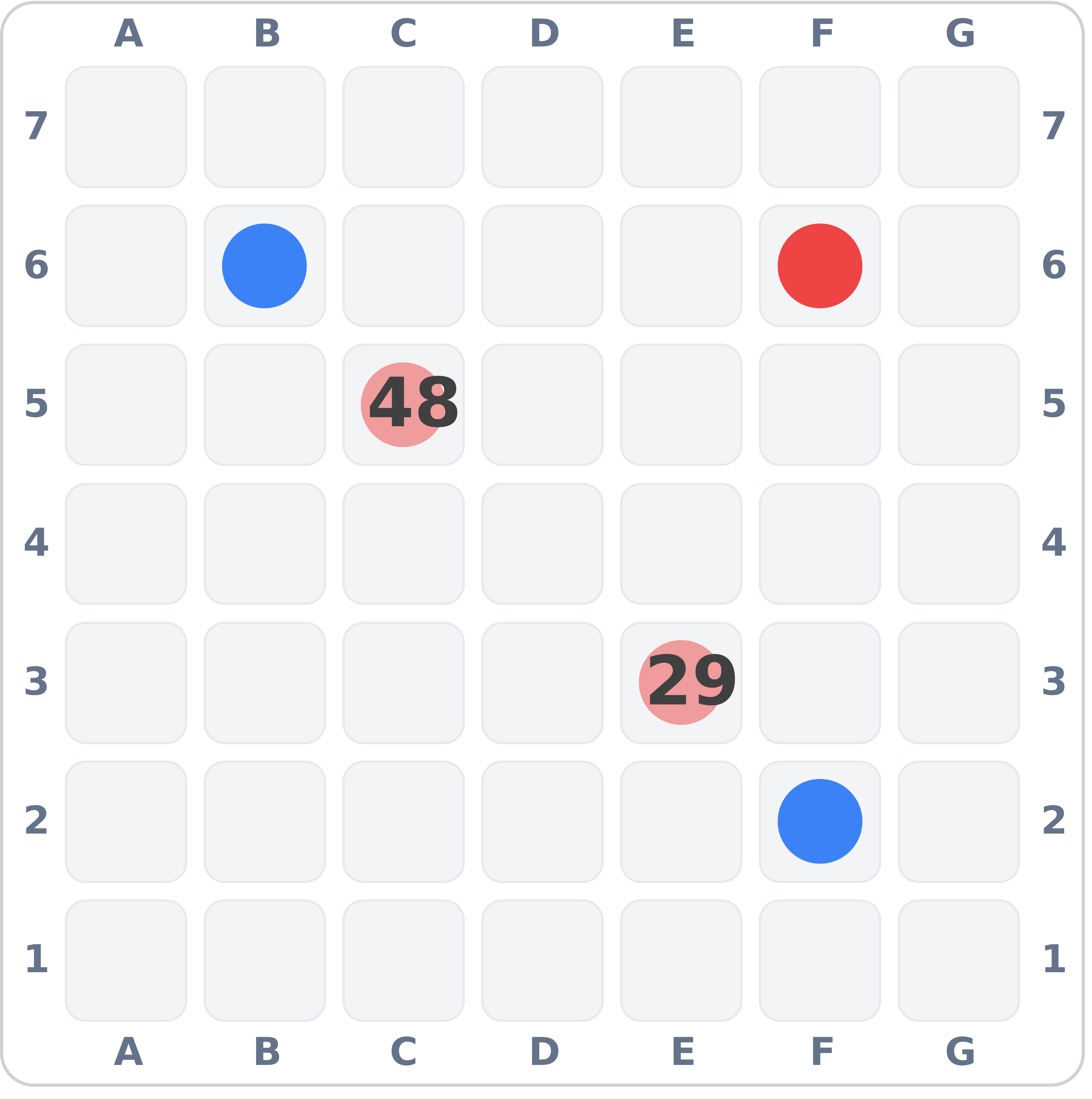}\label{fig:empty_step4}
    }
    \subfloat[Step 4\\53.4\%]{
        \includegraphics[width=0.15\textwidth]{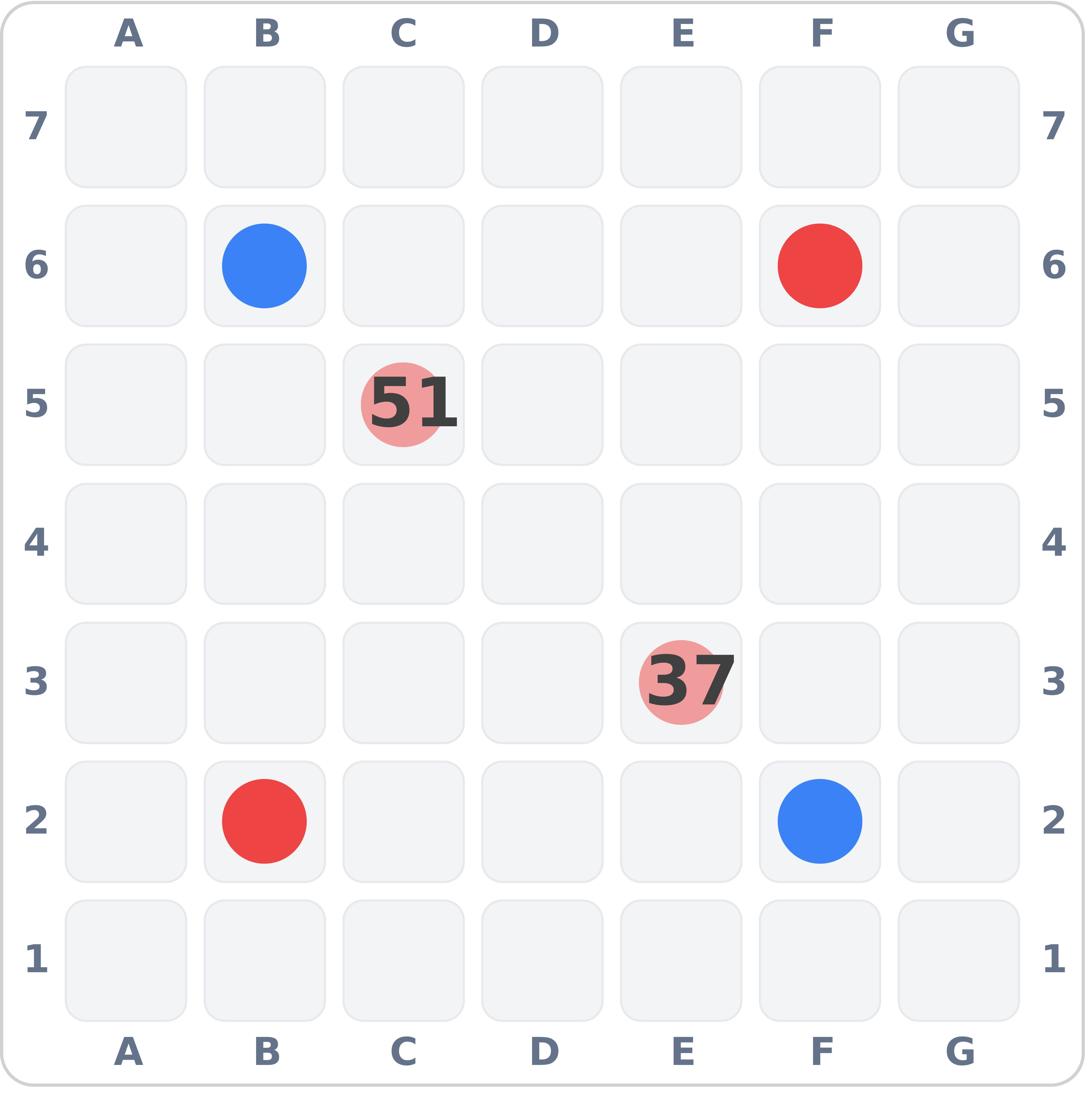}\label{fig:empty_step5}
    }
    \vrule width 0.8pt
    \subfloat[Step 0\\51.35\%]{
        \includegraphics[width=0.15\textwidth]{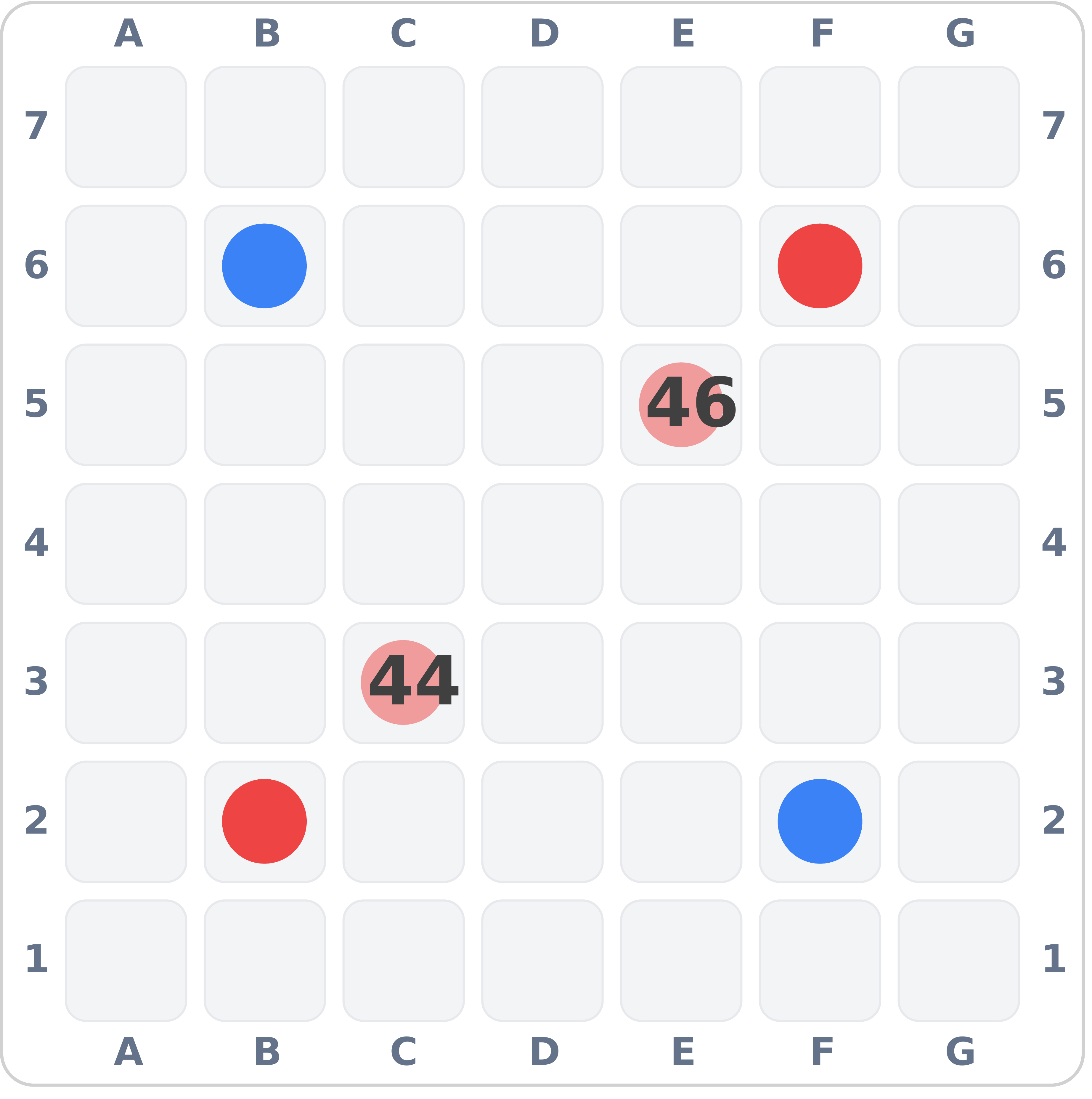}\label{fig:4-stone}
    }


    \caption{Win rate (from Red's perspective) and policy probabilities predicted by the policy and value networks.
    For the policyonly probabilities larger than 10\% are displayed.
    (a)–(e) use the empty mode model with the 4-stone opening; (f) uses the 4-stone mode model.}
    
    \label{fig:opening_winrate}
\end{figure}

Figures~\ref{fig:empty_step1}–\ref{fig:empty_step5} illustrate the evolution of win rate and policy probabilities as the 4-stone mode opening is progressively introduced from the empty mode.
As can be observed, the policy trained in empty mode does not converge toward the predefined 4-stone opening. 
At the initial state, it clearly prioritizes concentrating its probability mass in the middle of the board. 
More broadly, even when the opening is constructed incrementally, the agent systematically avoids all predefined locations, attributing near-zero probability to each of them.

In parallel, we observe that the empty mode exhibits a stronger first-player advantage, with a Red win rate of 55.95\% compared to 51.35\% in the 4-stone mode. 
When the four fixed stones of the 4-stone mode opening are progressively introduced, the model’s win rate shows an overall downward trend. 
Together, these results indicate that although the 4-stone opening introduced in \textit{The Devil's Plan} is not aligned with the strategy learned in empty mode, its symmetrical design leads to a more balanced game from the start.

\subsubsection{Analysis of Opening Distributions.}

\begin{figure}[ht] 
    \centering 
    \captionsetup[subfigure]{labelformat=empty, justification=centering} 

    \hspace{3.3em}
    \begin{minipage}{0.04\columnwidth} 
        \rotatebox{90}{\textbf{Empty Mode}} 
    \end{minipage} 
    \hspace{-3.5em}
    \begin{minipage}{0.94\columnwidth} 
        \centering 
        \subfloat[freq. = $5.8\%$]{ \includegraphics[width=0.2\textwidth]{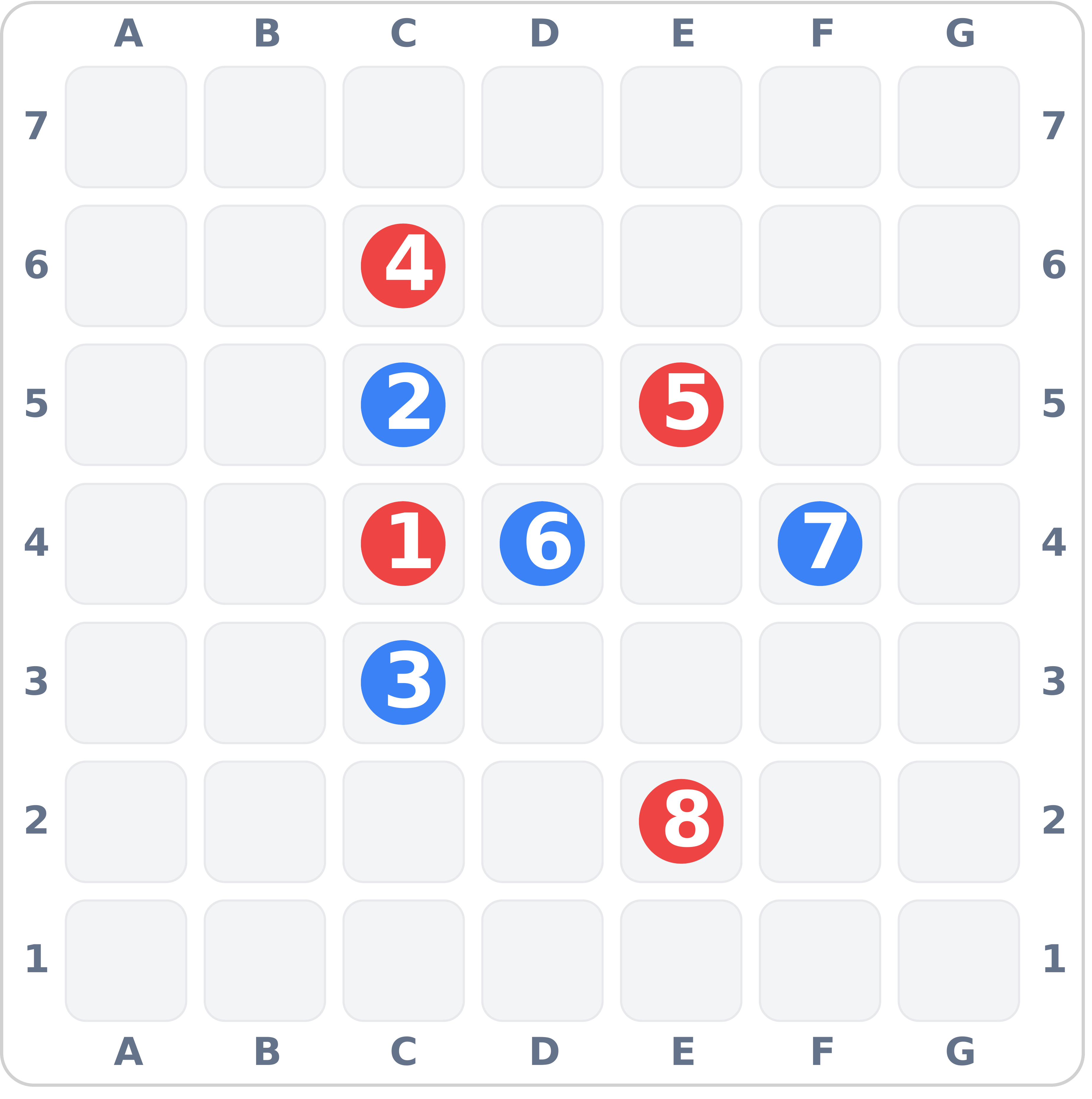} } 
        \subfloat[freq. = $5.0\%$]{ \includegraphics[width=0.2\textwidth]{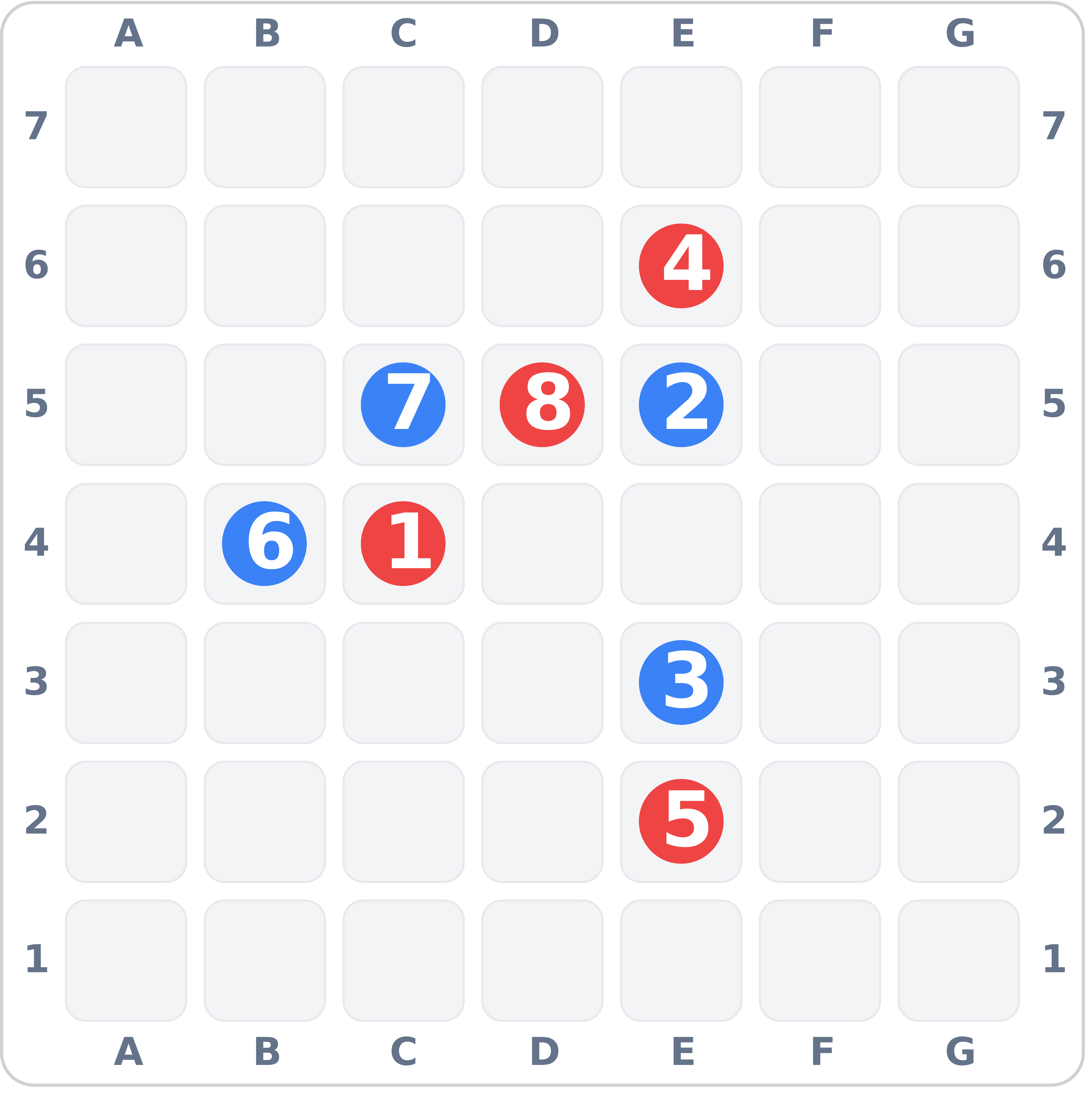} } 
        \subfloat[freq. = $4.5\%$]{ \includegraphics[width=0.2\textwidth]{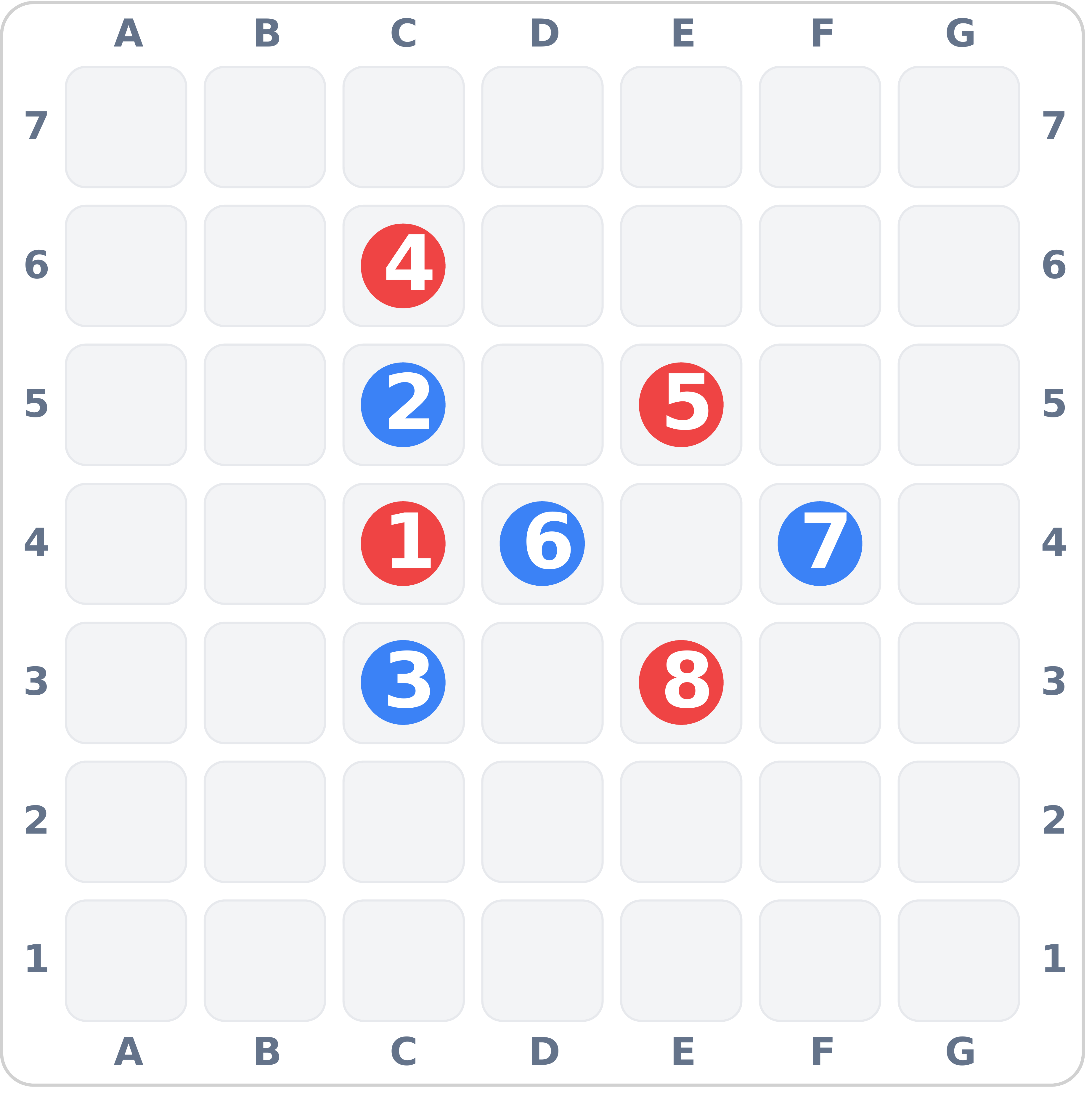} } 
        \subfloat[freq. = $4.1\%$]{ \includegraphics[width=0.2\textwidth]{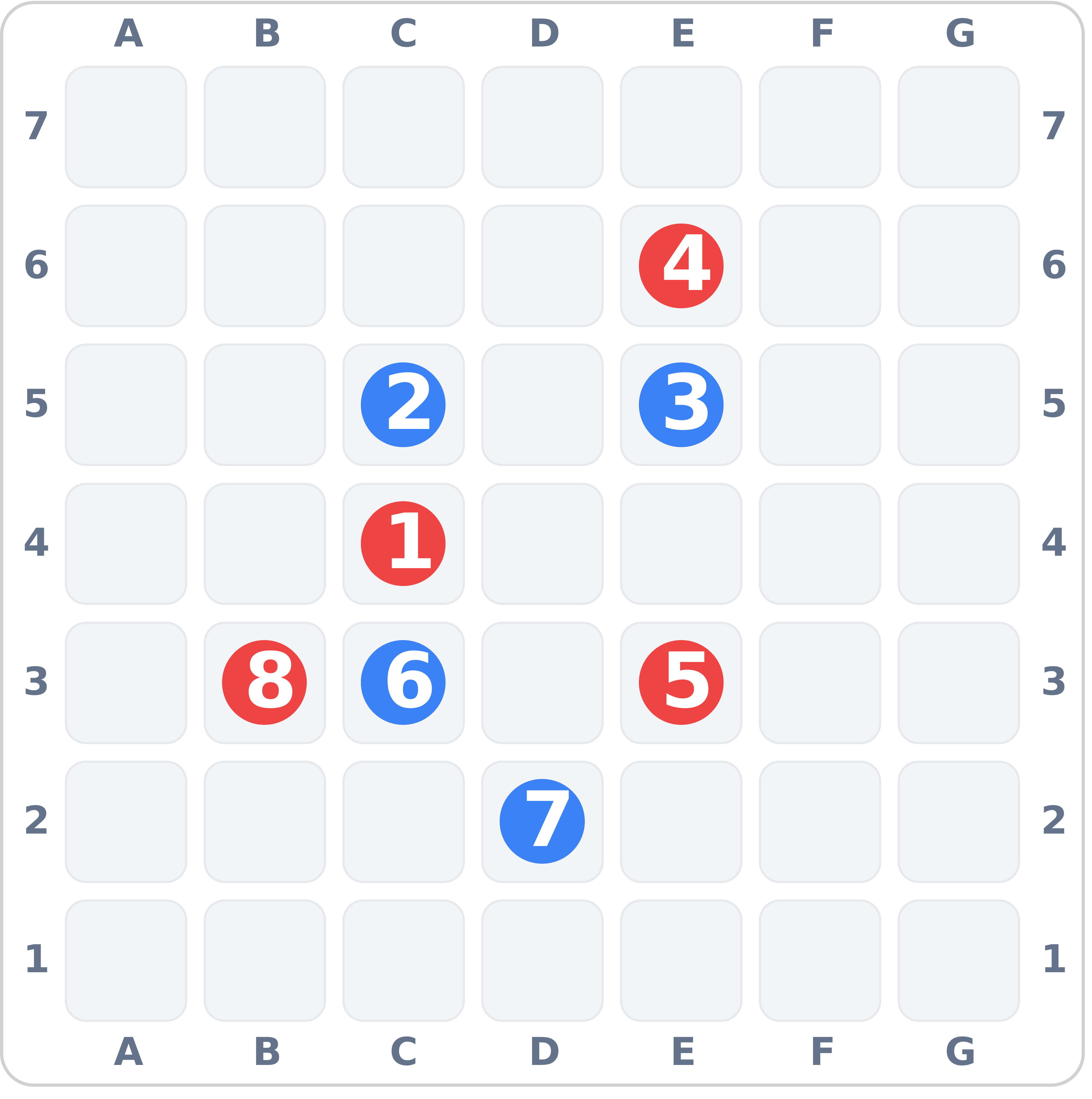} } 
    \end{minipage} 

    \hspace{3.3em}
    \begin{minipage}{0.04\columnwidth} 
        \rotatebox{90}{\textbf{4-Stone Mode}} 
    \end{minipage} 
    \hspace{-3.5em}
    \begin{minipage}{0.94\columnwidth} 
        \centering 
        \subfloat[freq. = $49.0\%$]{ \includegraphics[width=0.2\textwidth]{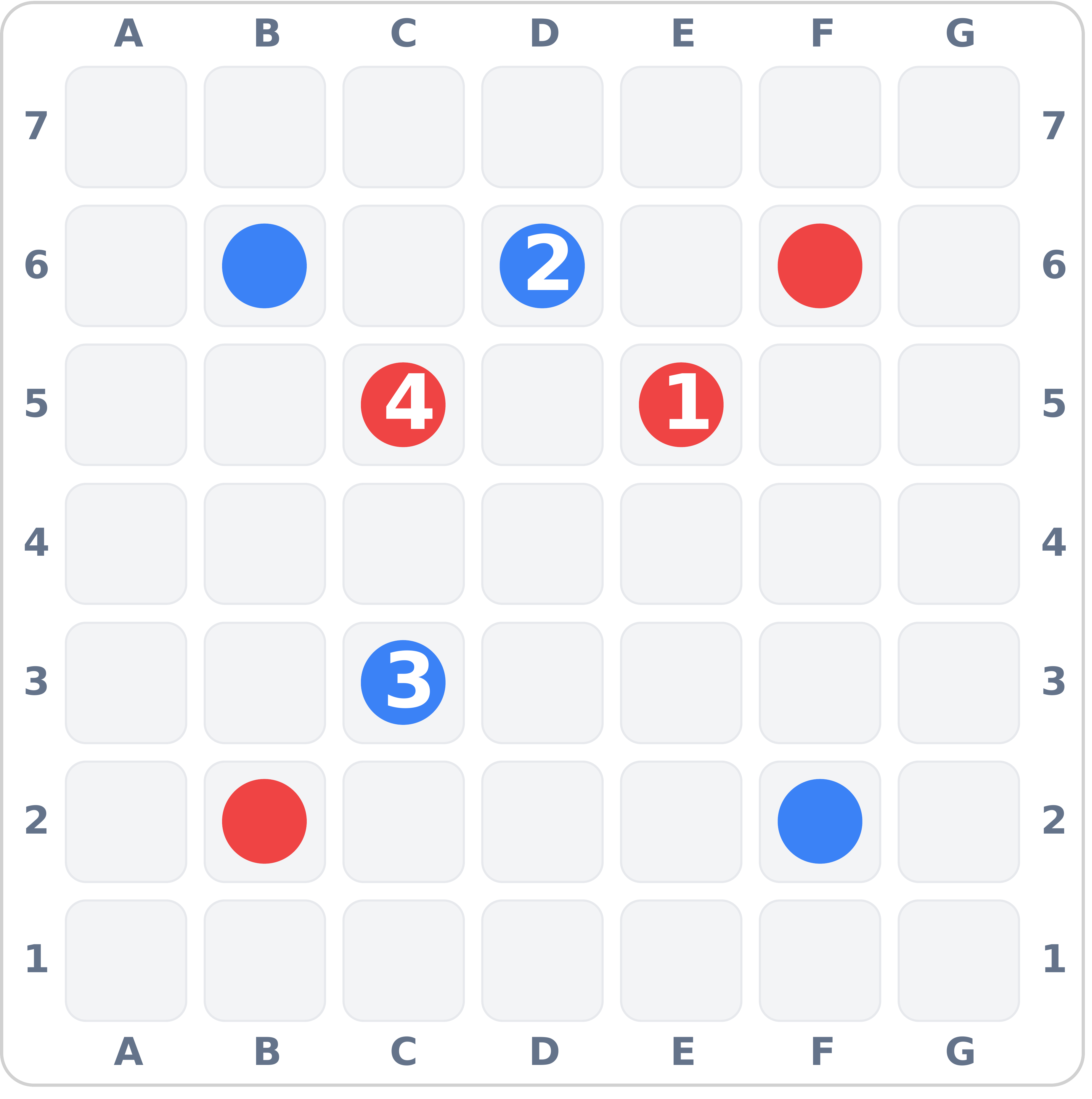} } 
        \subfloat[freq. = $21.0\%$]{ \includegraphics[width=0.2\textwidth]{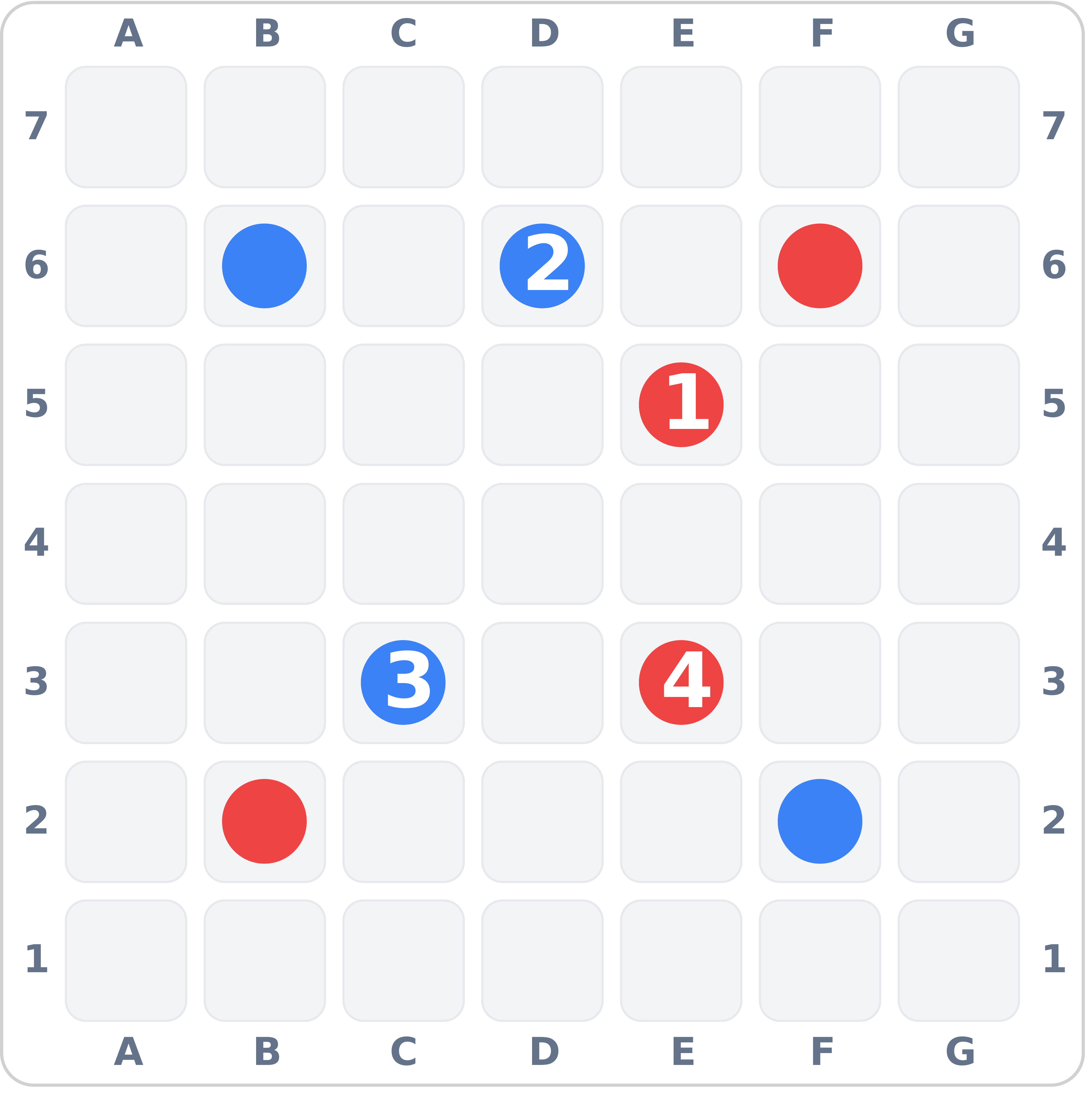} } 
        \subfloat[freq. = $15.1\%$]{ \includegraphics[width=0.2\textwidth]{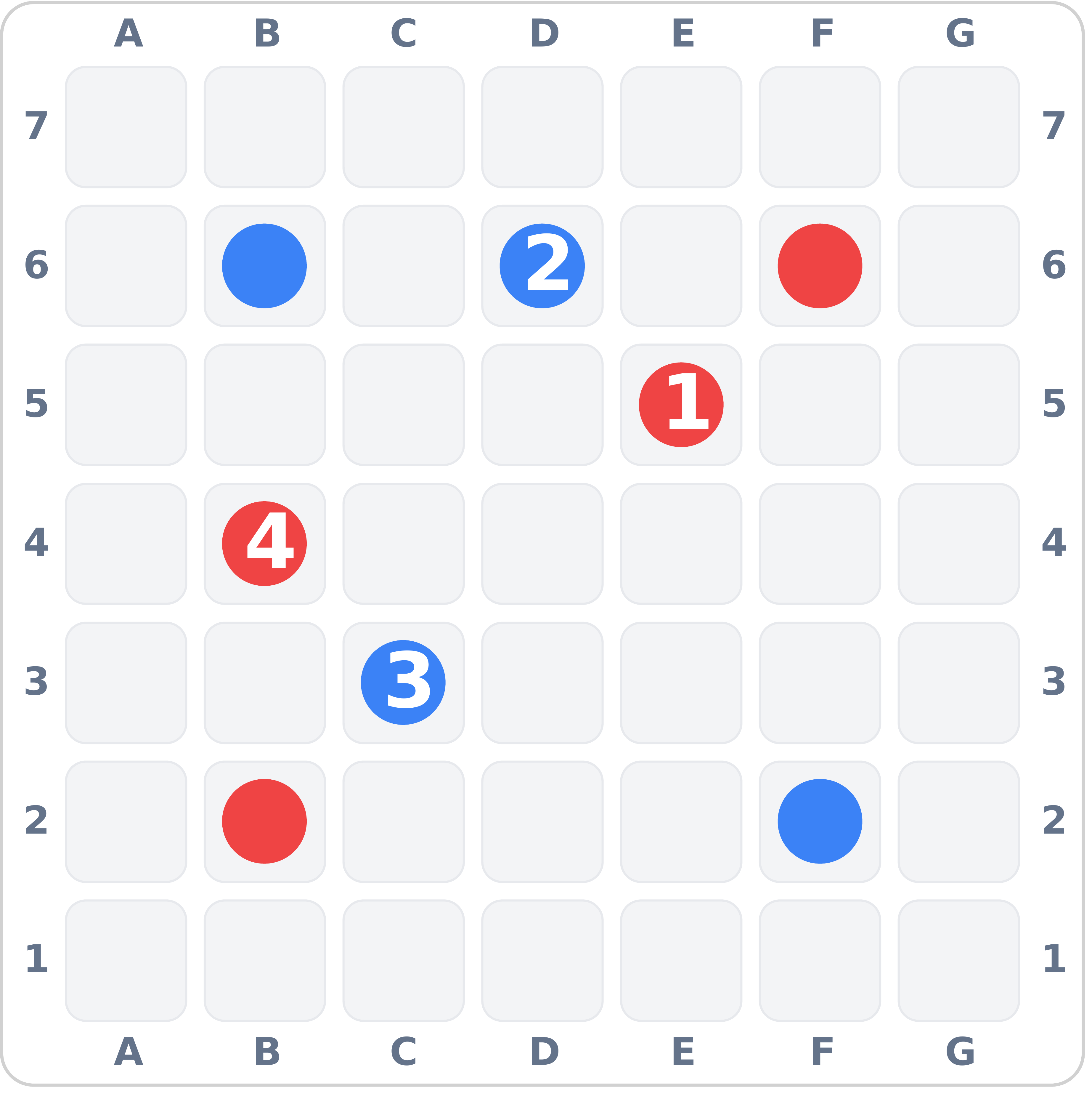} } 
        \subfloat[freq. = $1.8\%$]{ \includegraphics[width=0.2\textwidth]{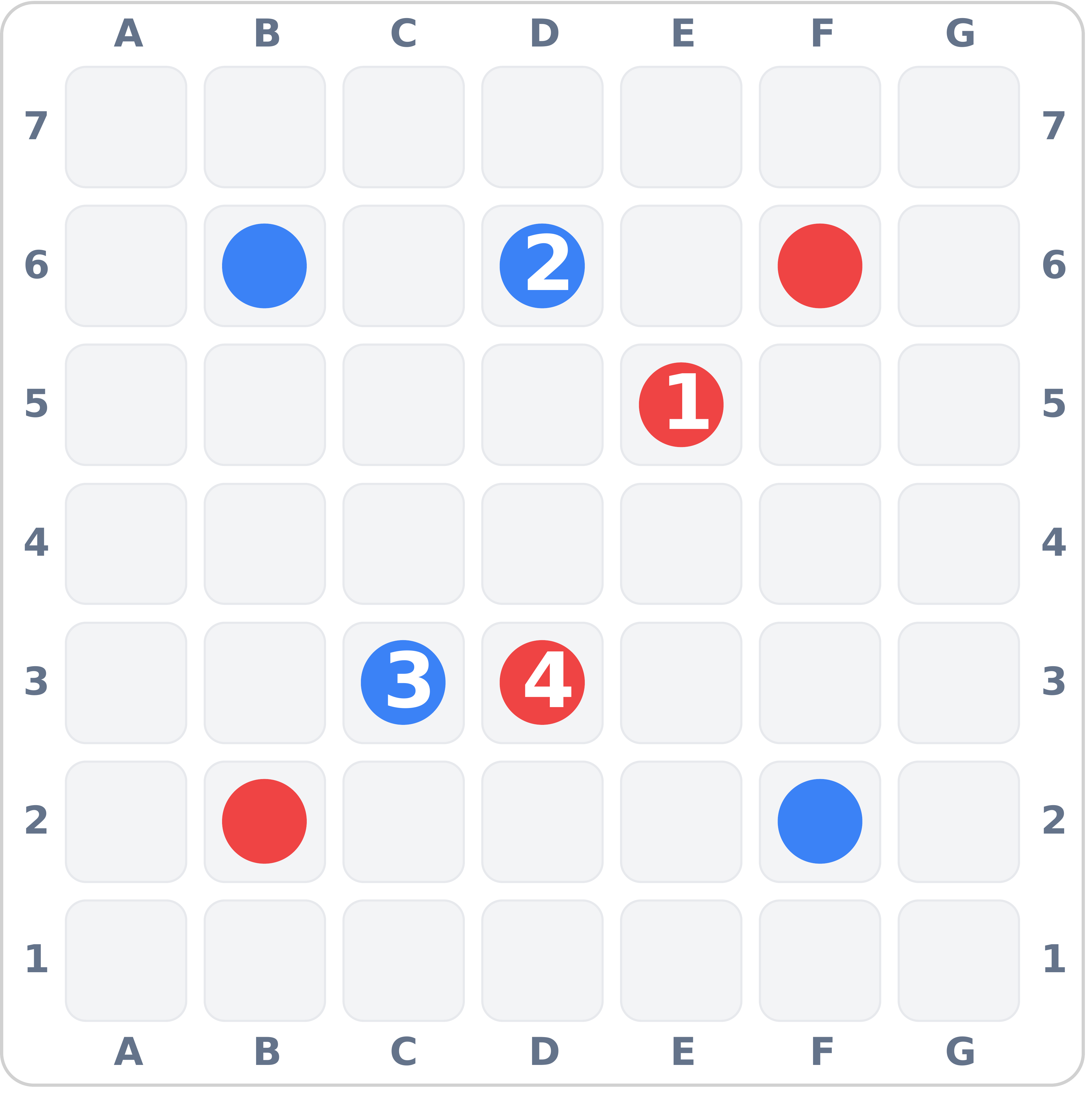} } 
    \end{minipage} 
    
    \caption{Top 4 openings in empty and 4-stone modes. 
    Percentages denote frequencies over 1,000 self-play games. 
    Numbers indicate move sequence.} 
    \label{fig:opening}
\end{figure}

Figure~\ref{fig:opening} illustrates the four most frequent opening states observed in empty mode and 4-stone mode, respectively.
In empty mode, openings exhibit high diversity with no dominant fixed patterns. 
Nonetheless, a consistent strategy emerges in which the initial moves are consistently concentrated in the center, as both players cluster around their opponent’s stones to restrict reachability.
In contrast, in 4-stone mode, openings converge toward a limited set of patterns, differing only in the fourth step.
Our analysis reveals strategic trends: Red uses the first move to consolidate and expand regional advantages, while Blue responds with two consecutive placements to counter Red’s influence. 
Finally, Red’s last placement is directed to a region that maintains containment over Blue’s forces.

\subsection{Strategic Analysis}

As WallGo is a newly introduced game, its strategies remain largely unexplored.
In this subsection, we analyze WallZero's self-play games in 4-stone mode together with insights from professional Go players to better understand WallGo's strategies.
We identify two core strategies: \textit{reachability control} and \textit{passing strategy}.

\subsubsection{Reachability Control.}
Reachability refers to the set of positions a player can access under the current board.
Since both stones and walls affect reachability, maintaining high reachability is essential in WallGo.
We present three cases illustrating how it guides movement and wall construction decisions.

\begin{figure}[h]
    \centering
    \captionsetup[subfigure]{justification=centering}
    
    \subfloat[]{
        \includegraphics[width=0.18\columnwidth]{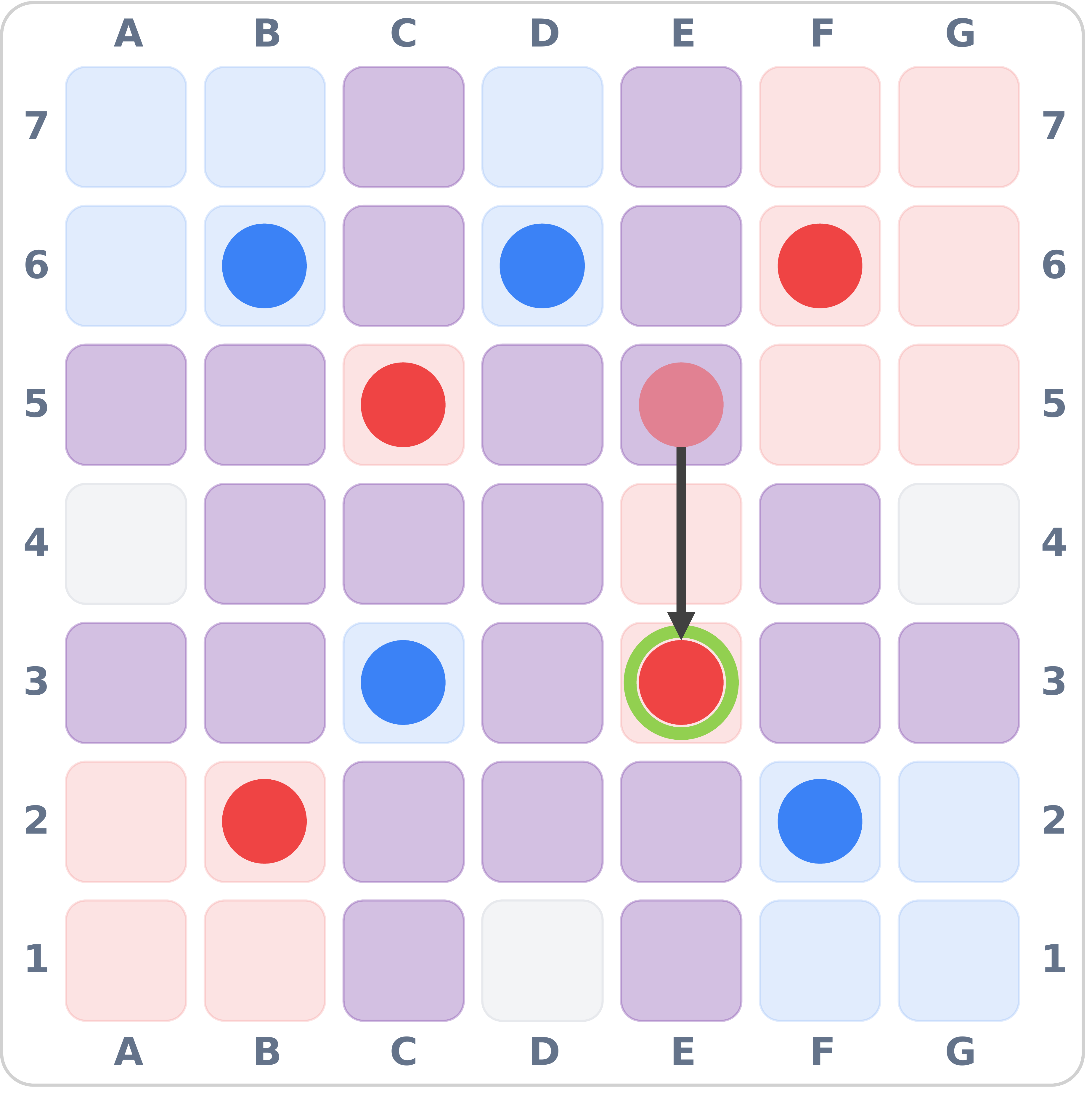}\label{fig:1_init}
    }
    \subfloat[$46\%$]{
        \includegraphics[width=0.18\columnwidth]{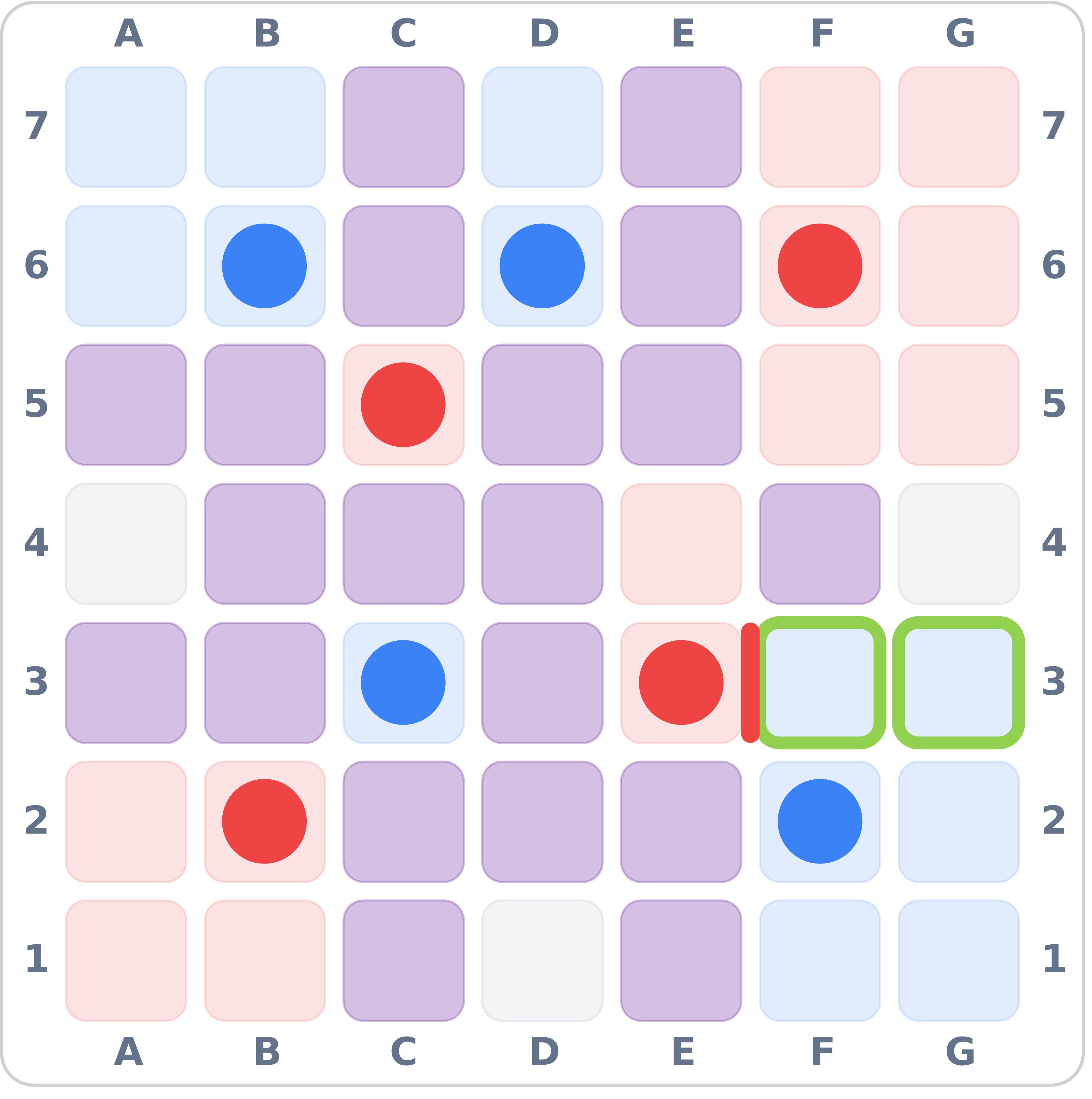}\label{fig:1_right}
    }
    \subfloat[$47\%$]{
        \includegraphics[width=0.18\columnwidth]{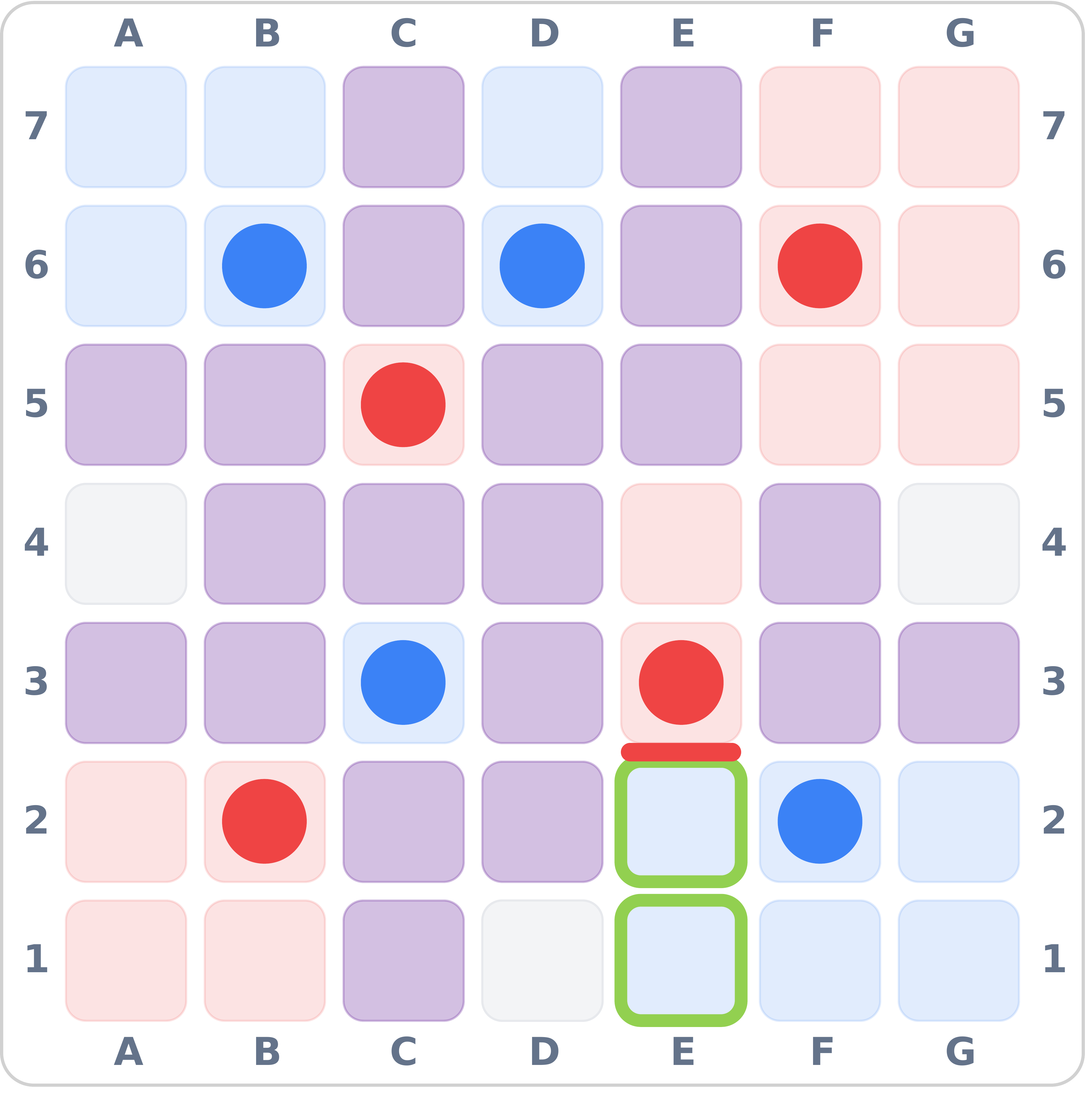}\label{fig:1_down}
    }
    \subfloat[$49.5\%$]{
        \includegraphics[width=0.18\columnwidth]{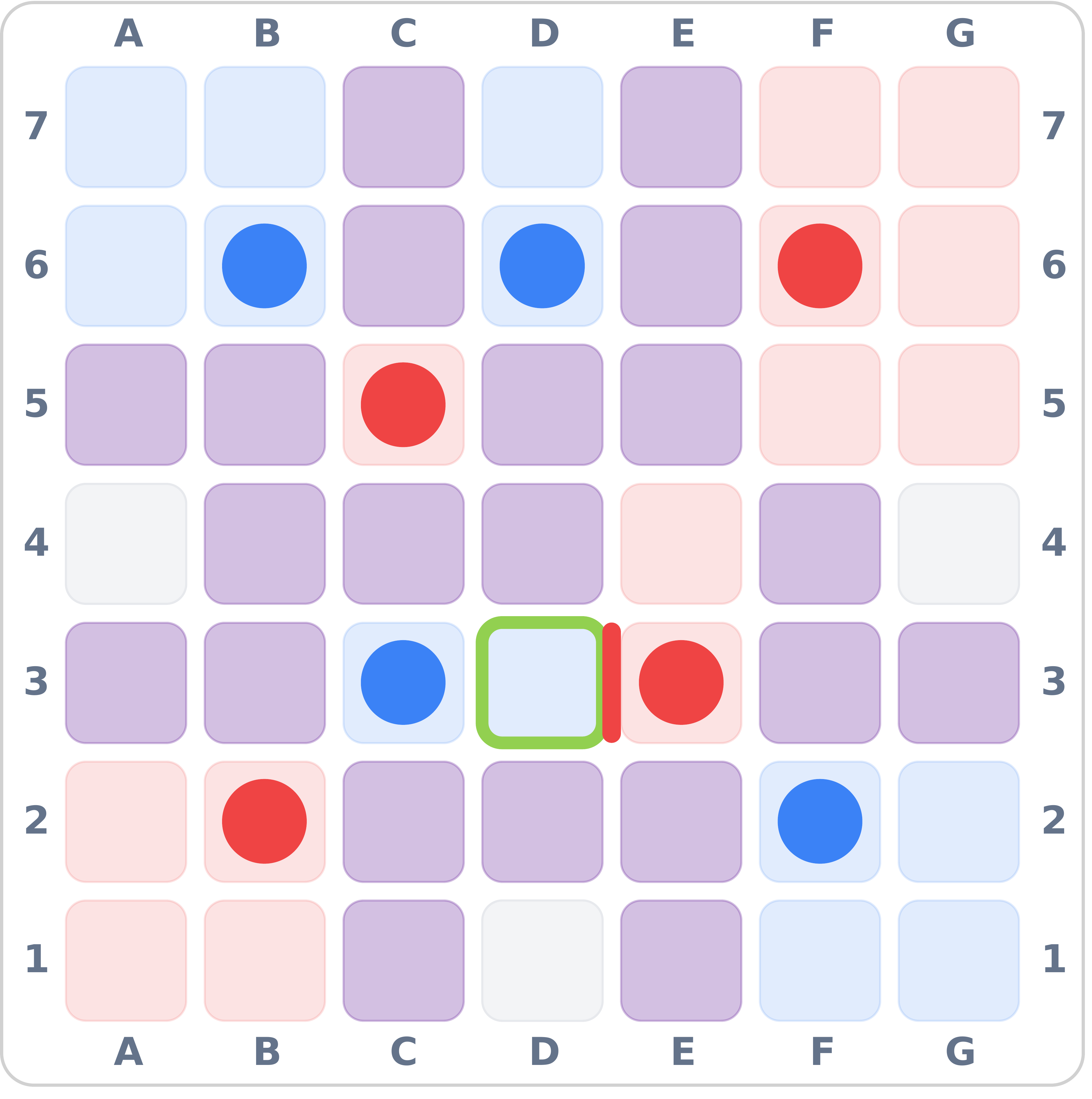}\label{fig:1_left}
    }
    \subfloat[$52.5\%$]{
        \includegraphics[width=0.18\columnwidth]{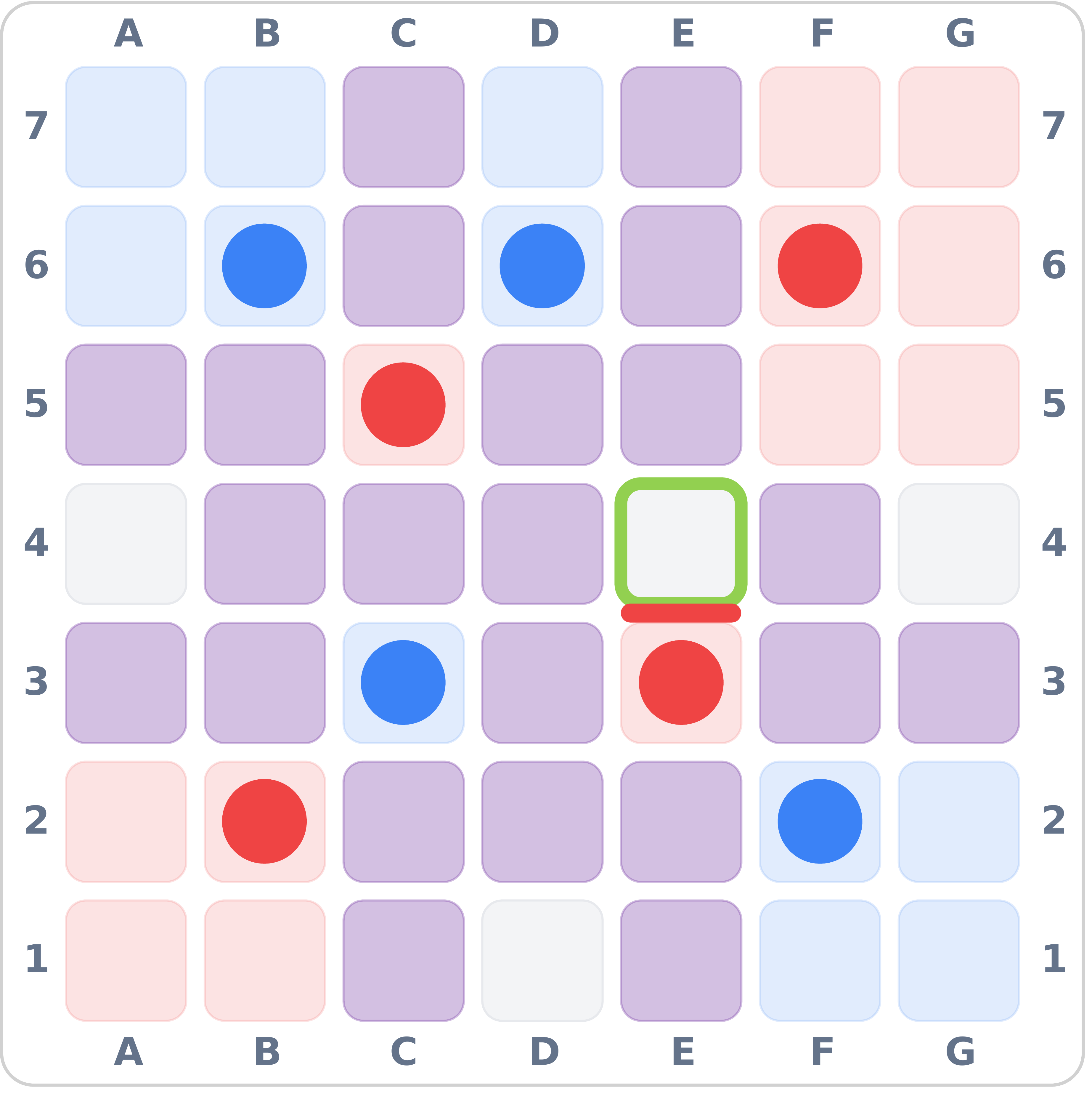}\label{fig:1_up}
    }
    
    \caption{Reachability Control -- Case I.
    (a) Initial position (Red moves E5 to E3).
    (b)-(e) Wall placed to the rightbelowleftand aboverespectively.
    Percentages denote the win rate from Red's perspective.
    Red and blue squares indicate exclusively reachable positions by the corresponding player; purple squares indicate positions reachable by both players.}
    \label{fig:reachable}
\end{figure}

\textbf{Case I (Figure~\ref{fig:reachable}).}
After Red moves from E5 to E3, different wall placements lead to distinct future reachability.
Placing the wall to the right or below loses two reachability (F3, G3 for right, and E2, E1 for below).
Although placing the wall to the left or above loses one reachability, D3 is exclusively reachable by Blue for the left placement, and E4 is unreachable to both players.
Therefore, placing the wall above is the most favorable move under WallZero's evaluation.

\begin{figure}[h]
    \centering
    \captionsetup[subfigure]{justification=centering}
    
    \subfloat[]{
        \includegraphics[width=0.18\columnwidth]{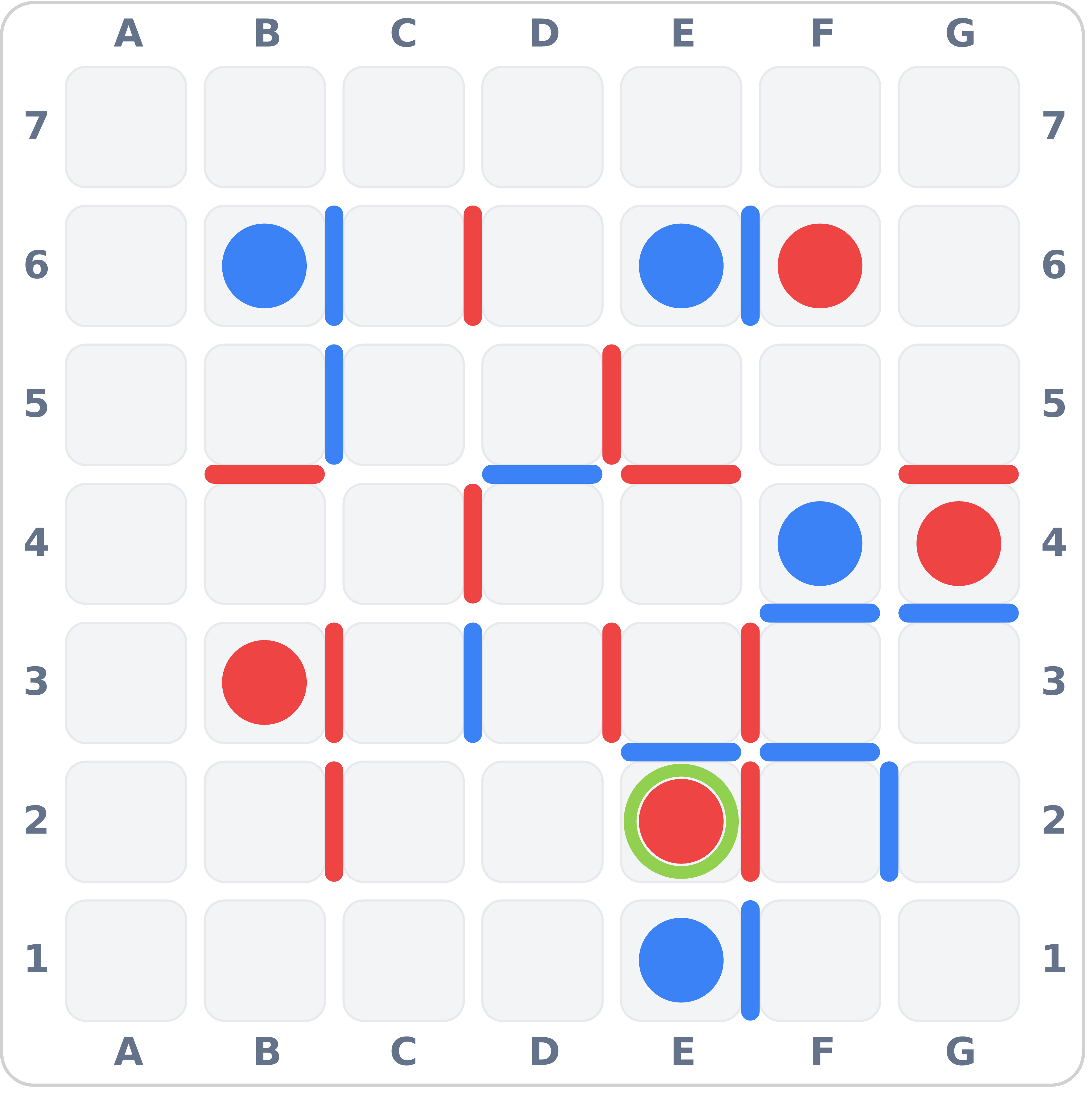}\label{fig:2_init}
    }
    \subfloat[]{
        \includegraphics[width=0.18\columnwidth]{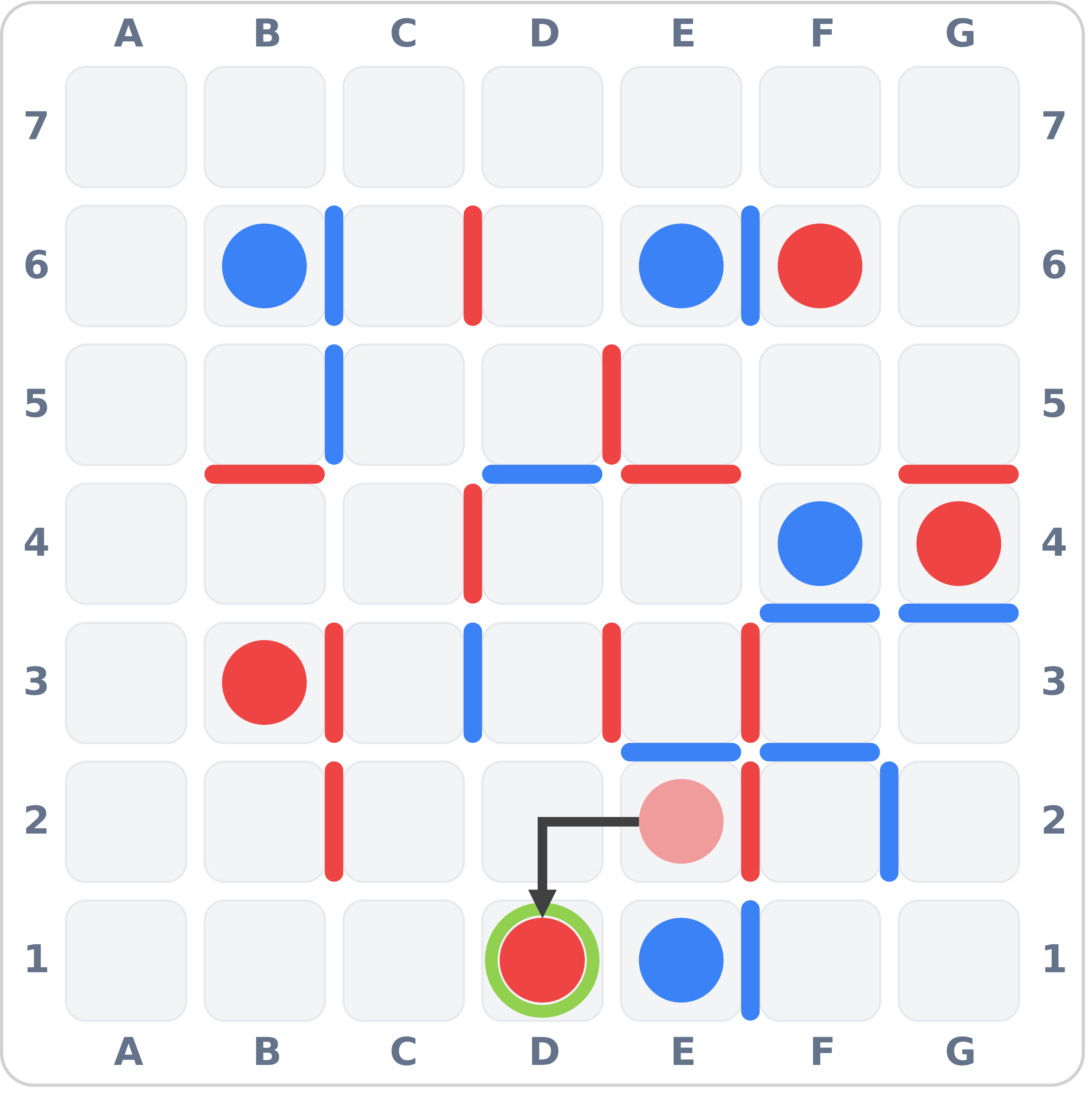}\label{fig:2_move}
    }
    \subfloat[$0.5\%$]{
        \includegraphics[width=0.18\columnwidth]{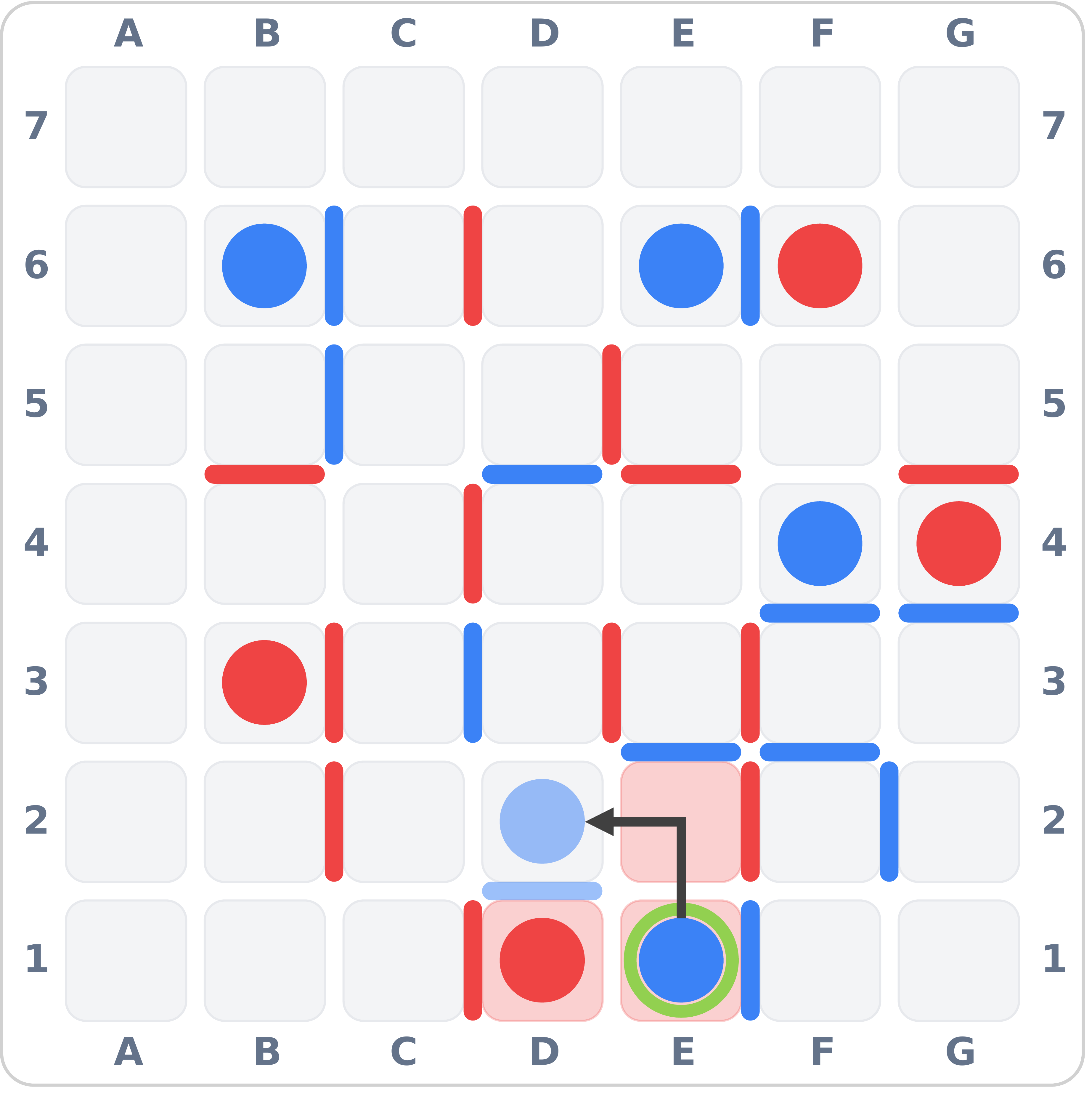}\label{fig:2_left}
    }
    \subfloat[$8\%$]{
        \includegraphics[width=0.18\columnwidth]{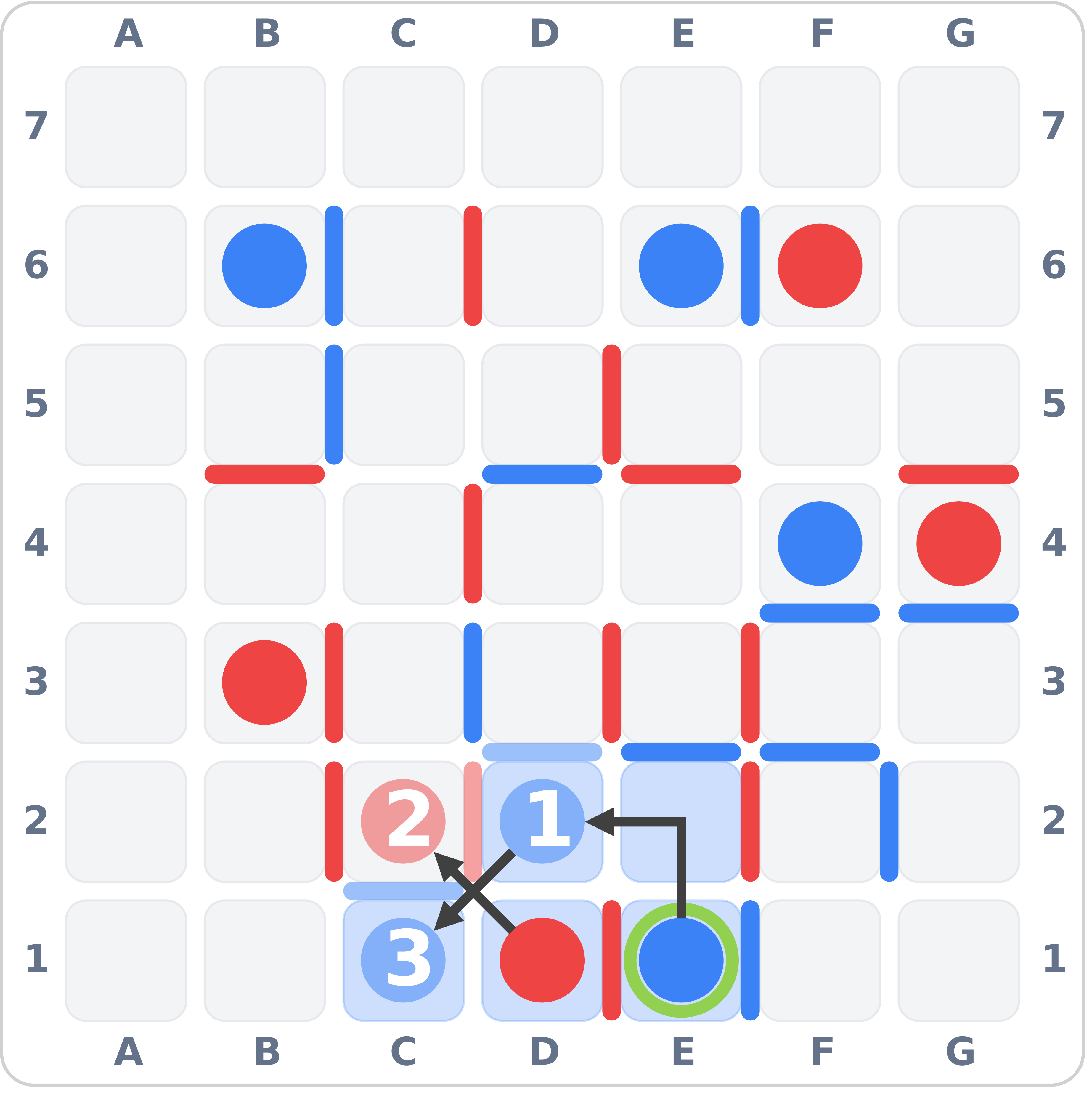}\label{fig:2_right}
    }
    \subfloat[$64.5\%$]{
        \includegraphics[width=0.18\columnwidth]{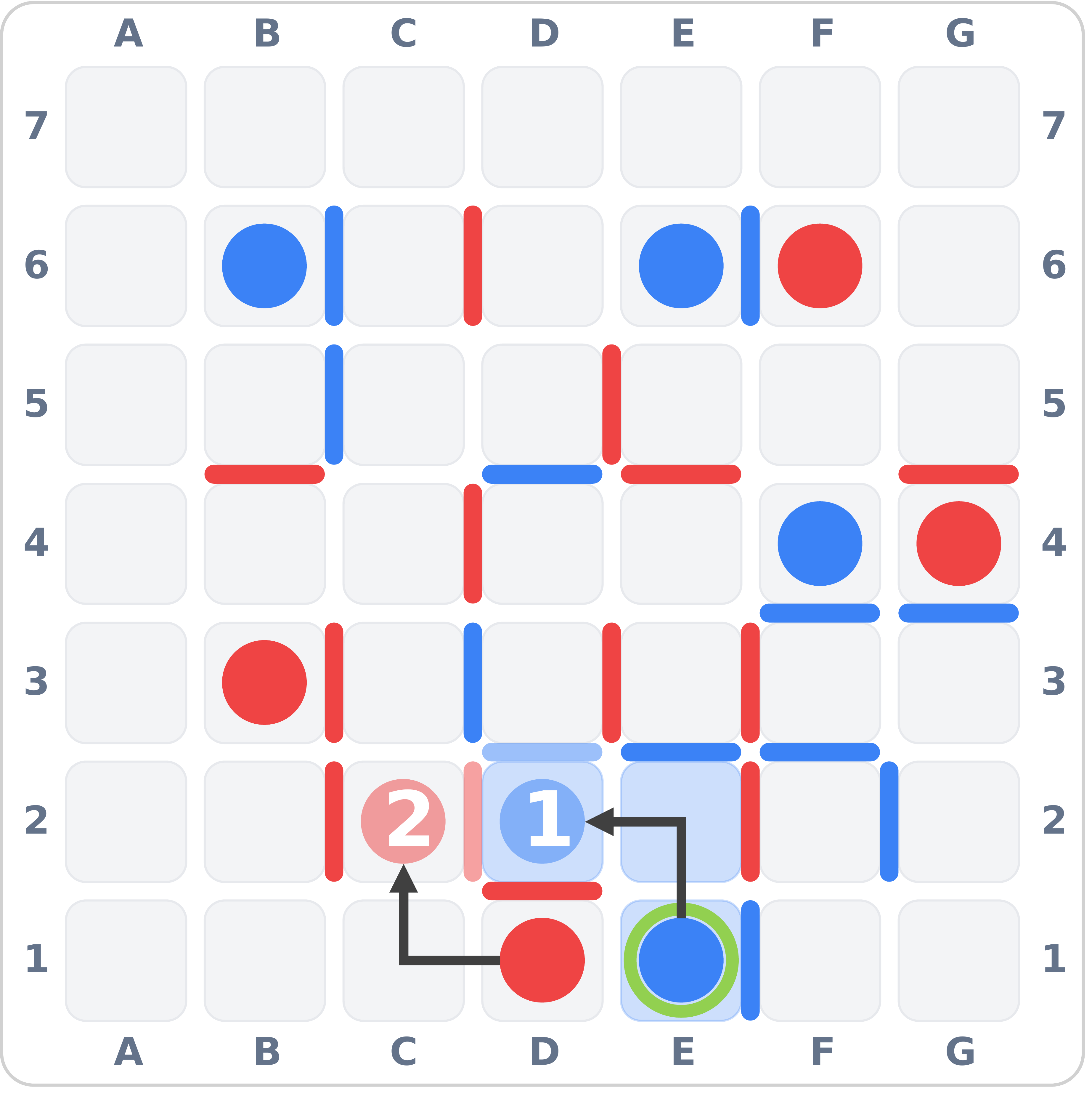}\label{fig:2_up}
    }
    
    \caption{Reachability Control -- Case II.
    (a) Initial position (Red to move).
    (b) Red moves E2 to D1.
    (c)-(e) Wall placed to the leftrightand aboverespectively.
    Percentages denote the win rate from Red's perspective.}
    \label{fig:reachable2}
\end{figure}

\textbf{Case II (Figure~\ref{fig:reachable2}).}
In this midgame position, Red at E2 and Blue at E1 compete for control of the lower region.
After Red moves from E2 to D1, three wall placements are available, each resulting in different reachable regions.
Building the wall to the left allows Blue to move from E1 to D2 and construct a wall below, reducing Red's territory region.
Building to the right improves upon the previous option, as Red is no longer blocked. However, this choice remains suboptimal as Blue can escape through D2 and C1, even if Red attempts to contain him by moving to C2. 
In contrast, placing the wall above preserves Red's reachability while limiting Blue's expansion.
If Blue moves to E2, Red can respond by moving to C2 and building a right wall, completely blocking Blue's leftward reachability.
This move achieves the highest value by WallZero.
Notably, the Red stone at D1 itself acts as a temporary barrier against Blue at E1.
This case highlights a key technique in WallGo: stones can serve as implicit walls, reducing the opponent's reachable region while preserving flexibility.

\begin{figure}[h]
    \centering
    \captionsetup[subfigure]{justification=centering}
    
    \subfloat[]{
        \includegraphics[width=0.18\columnwidth]{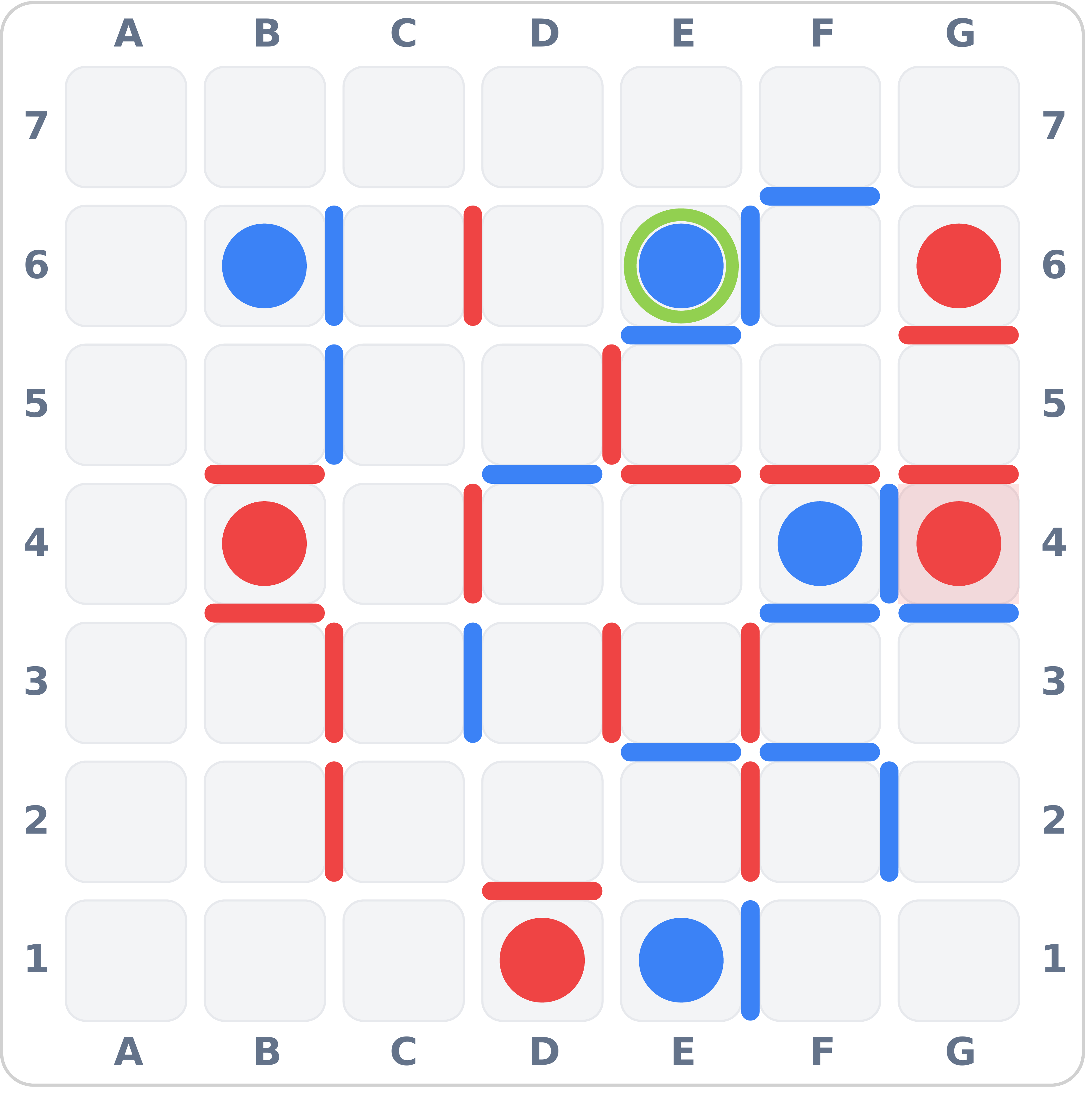}\label{fig:3_init}
    }
    \subfloat[$17.5\%$]{
        \includegraphics[width=0.18\columnwidth]{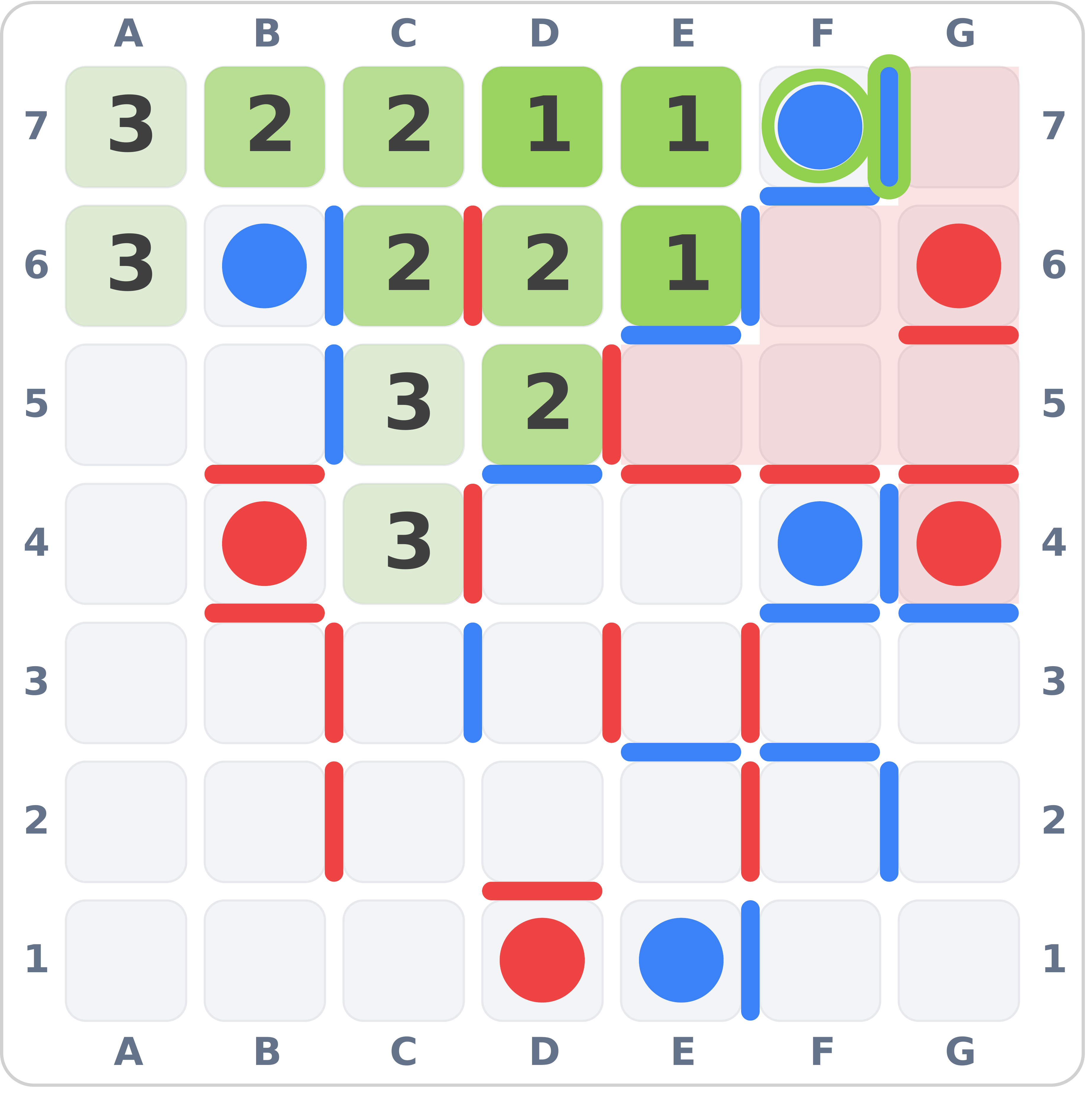}\label{fig:3_F1}
    }
    \subfloat[$55\%$]{
        \includegraphics[width=0.18\columnwidth]{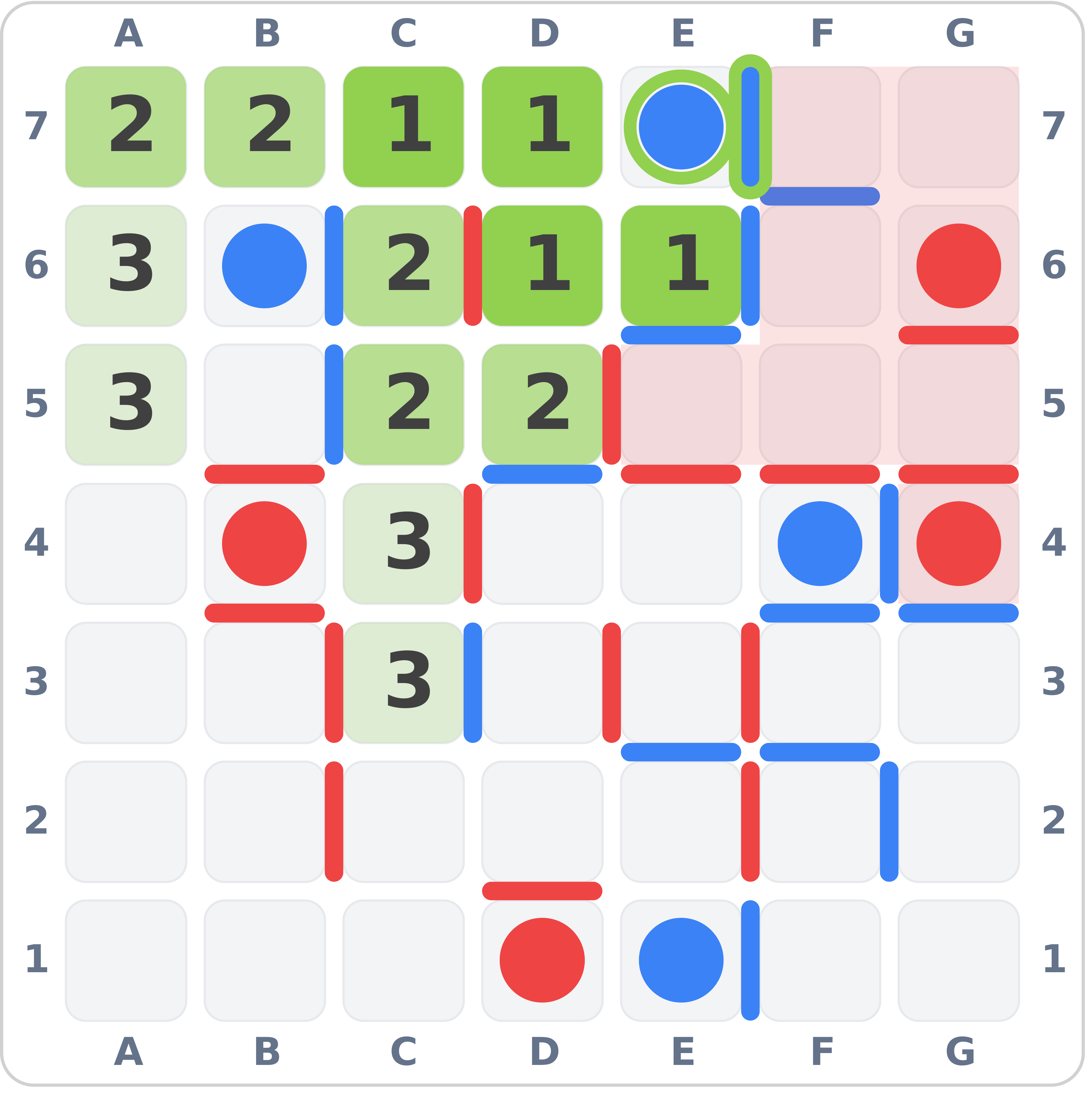}\label{fig:3_E1}
    }
    \subfloat[Red Win \\ (R:22, B:21)]{
        \includegraphics[width=0.18\columnwidth]{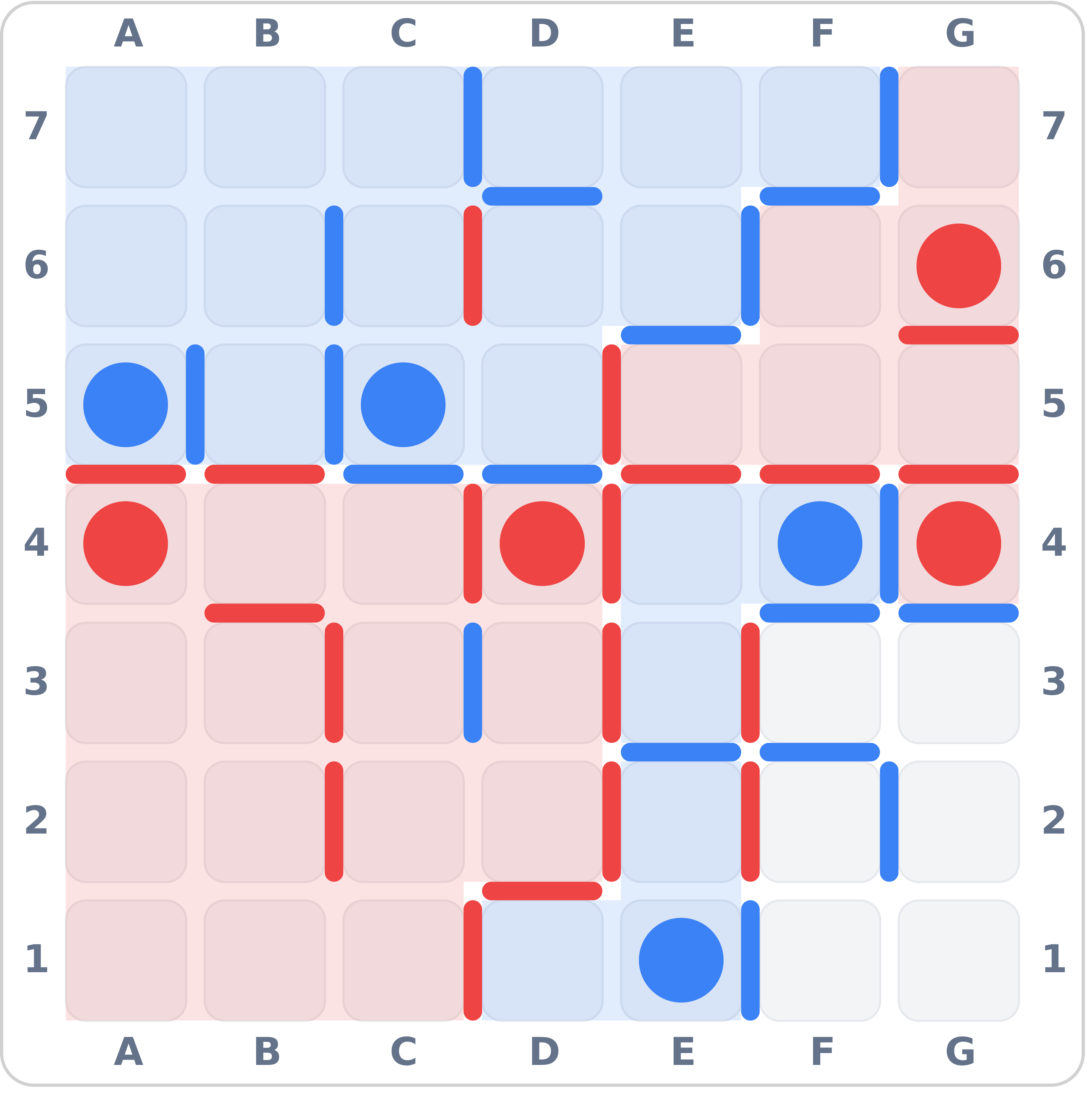}\label{fig:3_F1_final}
    }
    \subfloat[Blue Win \\ (R:21, B:22)]{
        \includegraphics[width=0.18\columnwidth]{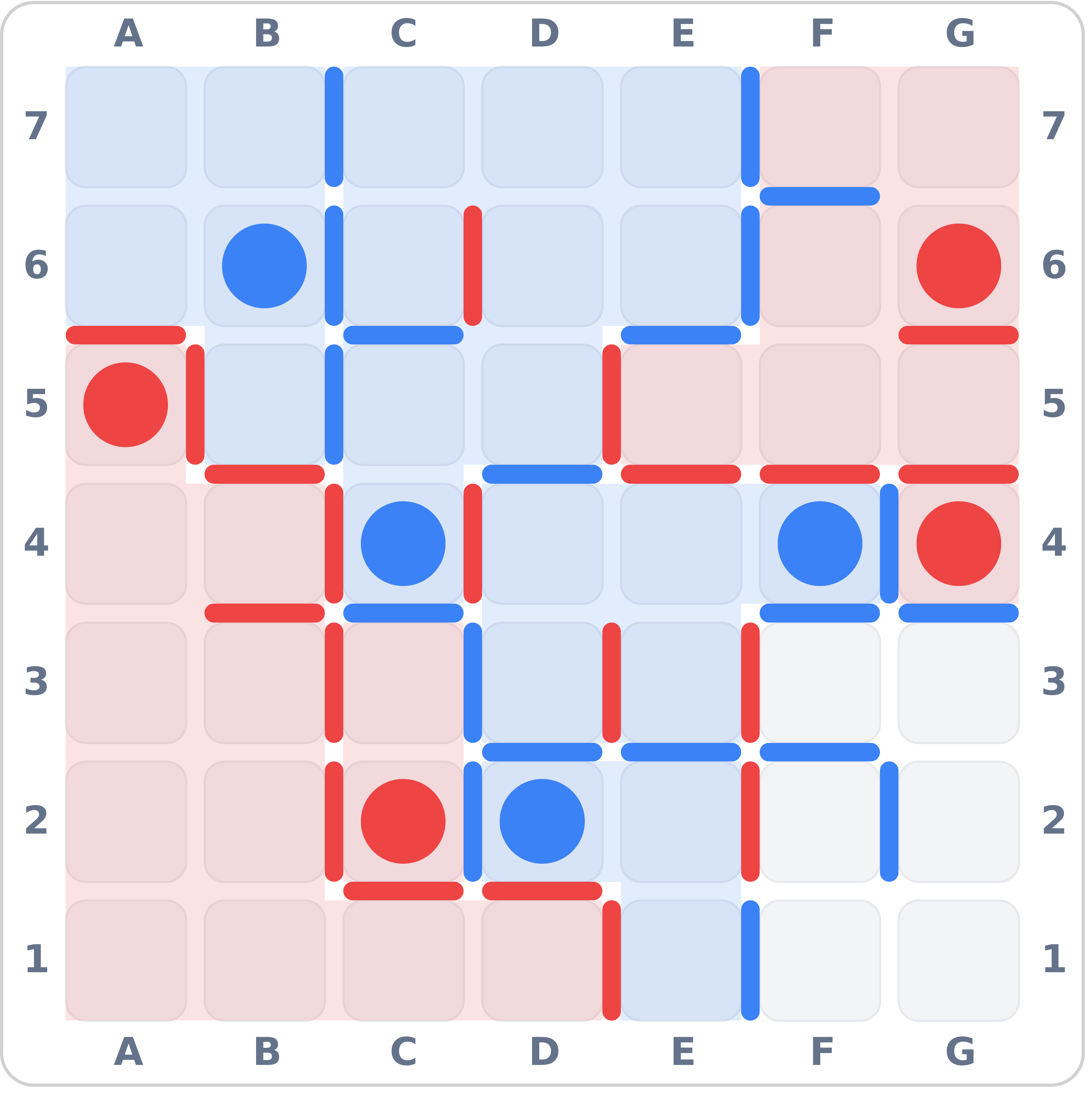}\label{fig:3_E1_final}
    }
    
    \caption{Reachability Control -- Case III. 
    (a) Initial position (Blue to move).
    (b)-(c) Blue moves to F7 and E7, respectively. Numbers on the board represent the number of steps required for the stone at F7 and E7 to reach each position.
    (d)-(e) Final outcomes under continued play. 
    Percentages denote the win rate from Blue's perspective.
    (R:x, B:y) indicate territory scores.}
    \label{fig:territory}
\end{figure}

\textbf{Case III (Figure~\ref{fig:territory}).}
This case presents an advanced technique in which a player sacrifices territory to preserve reachability.
In this position, Blue selects the stone at E6 to move.
An intuitive choice is to move to F7 (Figure~\ref{fig:3_F1}), minimizing the area that Red at G6 can enclose.
However, WallZero instead moves to E7 (Figure~\ref{fig:3_E1}), allowing Red to gain one additional point.
Although counterintuitive, this move is critical to the outcome.
Moving to E7 maintains control over the central battle around the Red stone at B4.
From F7, reaching C5 requires three steps, whereas from E7 it requires only two.
We further use WallZero to continue the game from these two moves.
Forcing Blue to move to F7 results in a one-point Red win (Figure~\ref{fig:3_F1_final}), while choosing E7 leads to a one-point Blue win (Figure~\ref{fig:3_E1_final}).
This example highlights that reachability can sometimes be more important than immediate territorial gain.

\subsubsection{Passing Strategy.}
In WallGo, a player must move a stone and build a wall on every turn, often restricting reachability for both players.
As a result, in certain positions, the ability to pass would be strategically advantageous.

\begin{figure}[h]
    \centering
    \captionsetup[subfigure]{justification=centering}
    
    \subfloat[Step 53\\(R:13, B:14)]{
        \includegraphics[width=0.18\columnwidth]{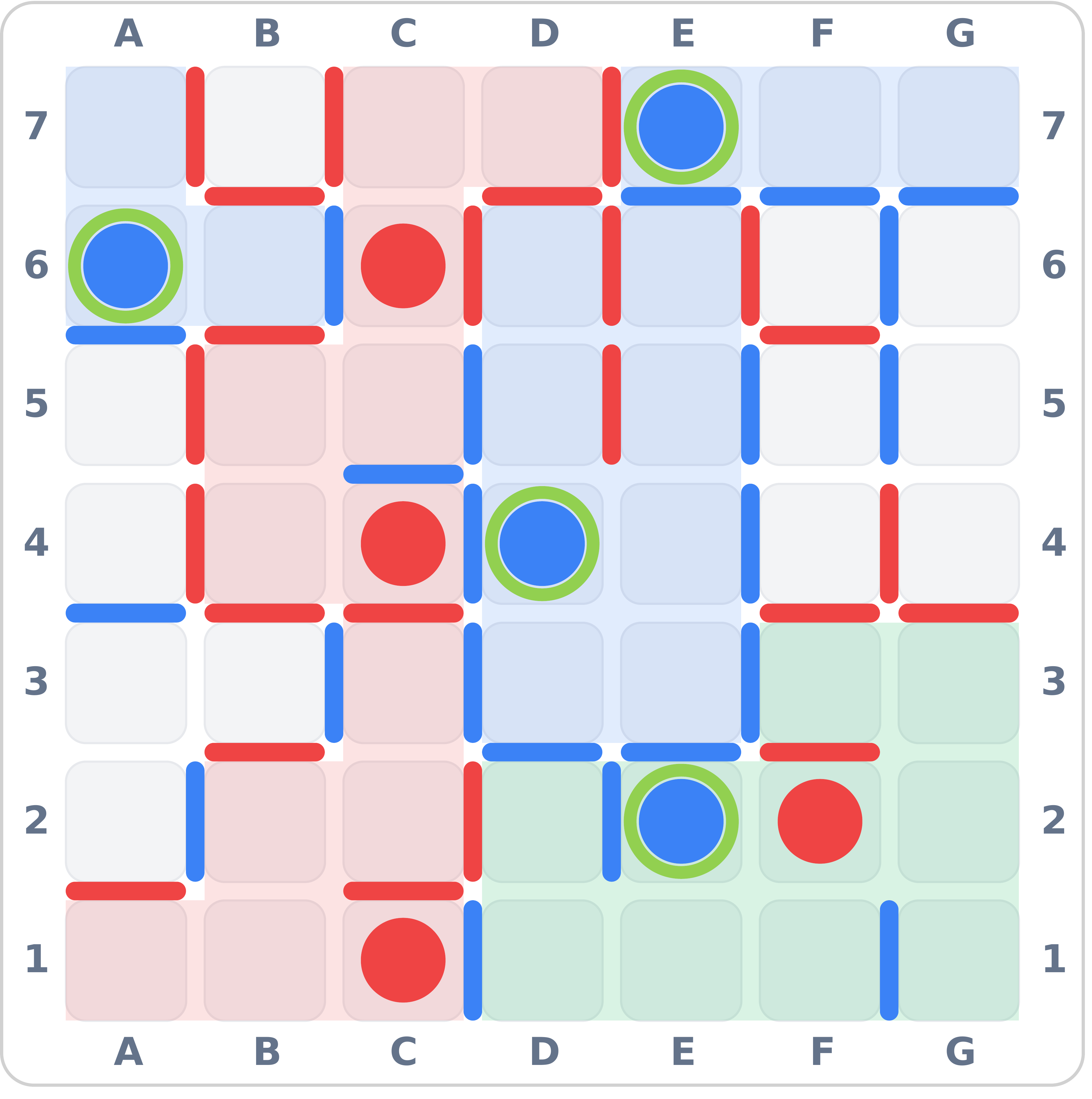}\label{fig:step53}
    }
    \subfloat[Red Win\\(R:20, B:17)]{
        \includegraphics[width=0.18\columnwidth]{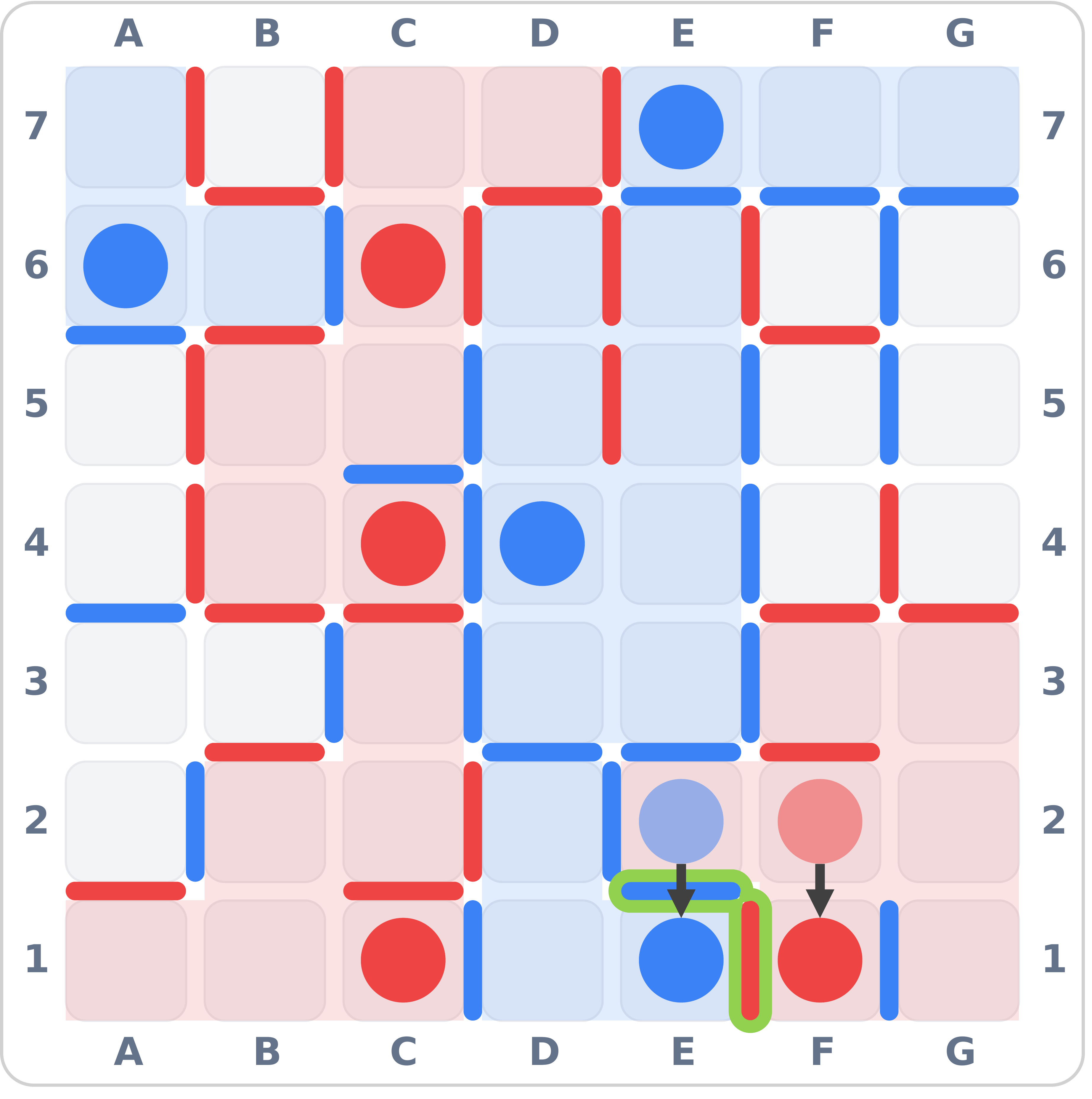}\label{fig:step53_blue_final}
    }
    \subfloat[Blue Win\\(R:18, B:19)]{
        \includegraphics[width=0.18\columnwidth]{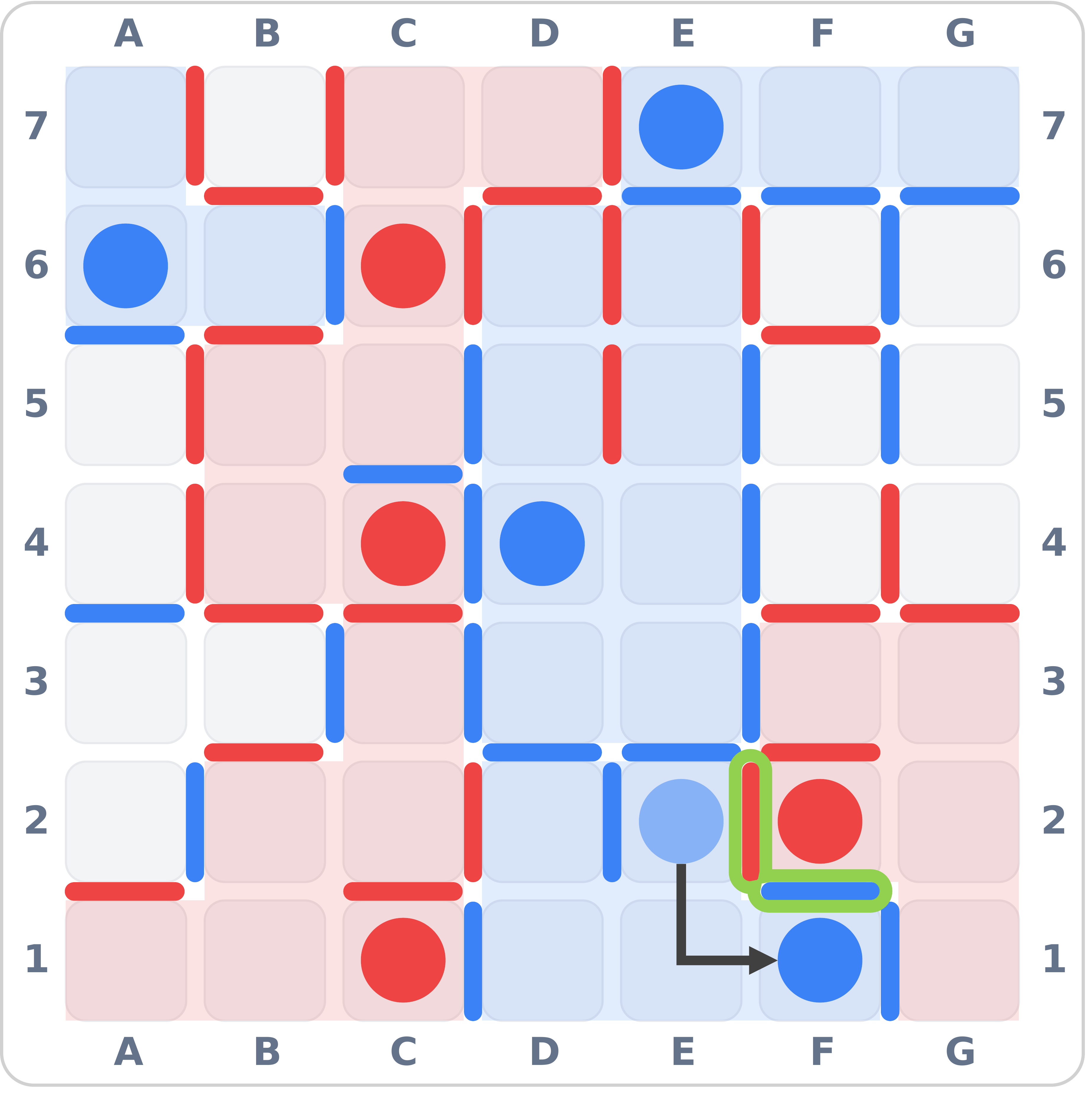}\label{fig:step53_red_final}
    }
    \subfloat[Step 54\\(R:13, B:14)]{
        \includegraphics[width=0.18\columnwidth]{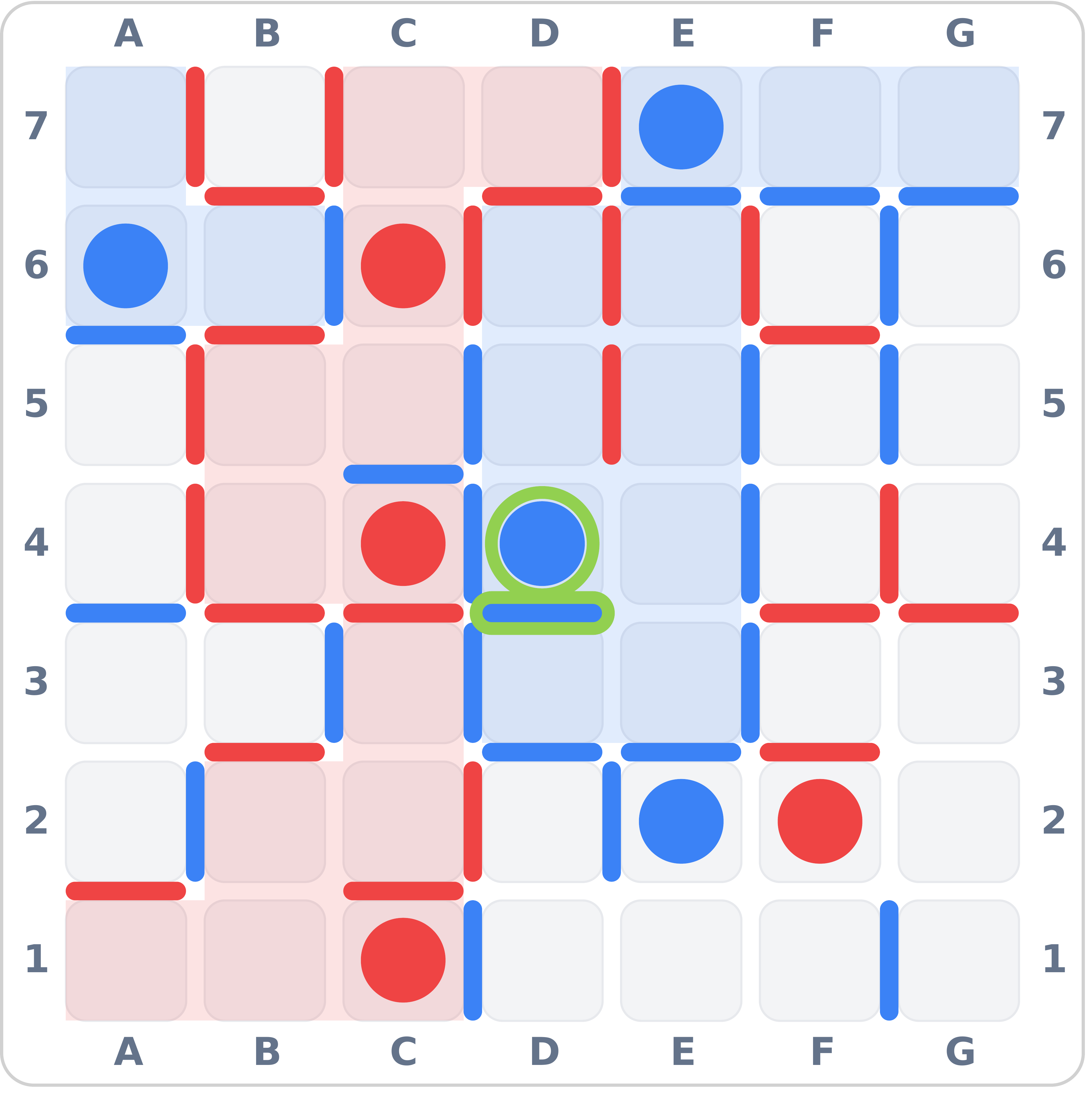}\label{fig:step54}
    }
    \hfill
    \subfloat[Step 55\\(R:12, B:14)]{
        \includegraphics[width=0.18\columnwidth]{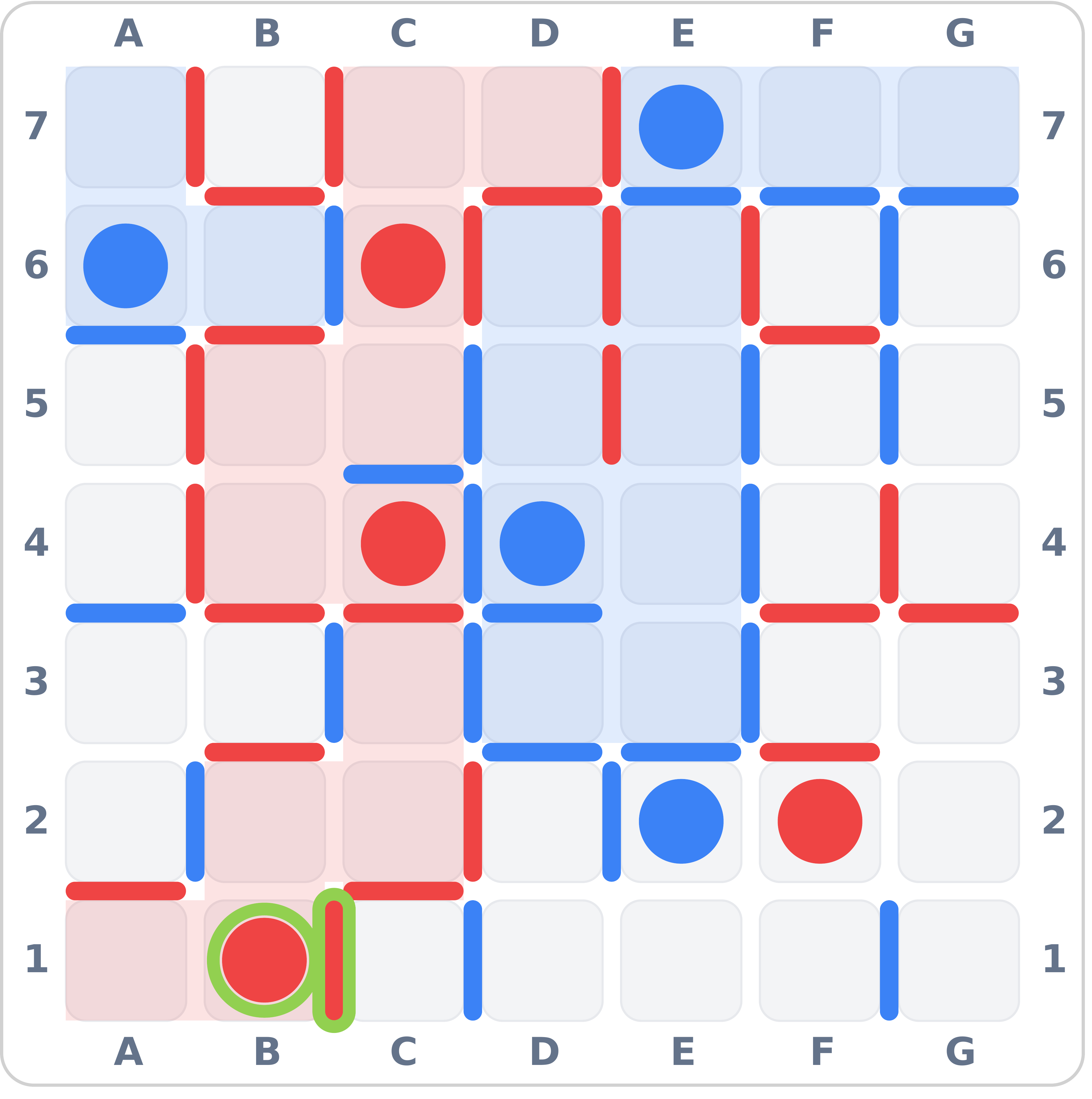}\label{fig:step55}
    }
    \subfloat[Step 76\\(R:3, B:3)]{
        \includegraphics[width=0.18\columnwidth]{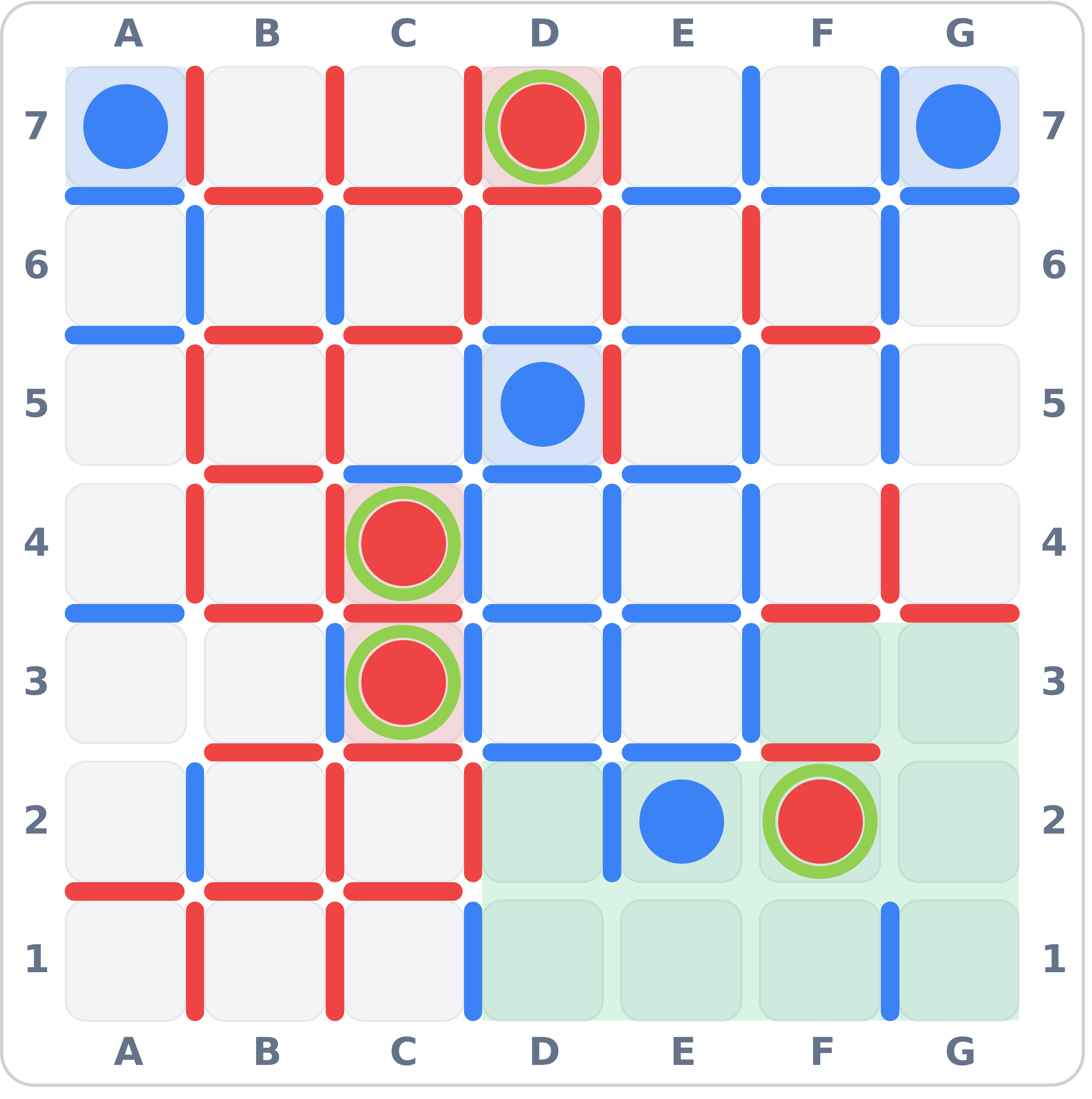}\label{fig:step76}
    }
    \subfloat[Step 77\\(R:3, B:3)]{
        \includegraphics[width=0.18\columnwidth]{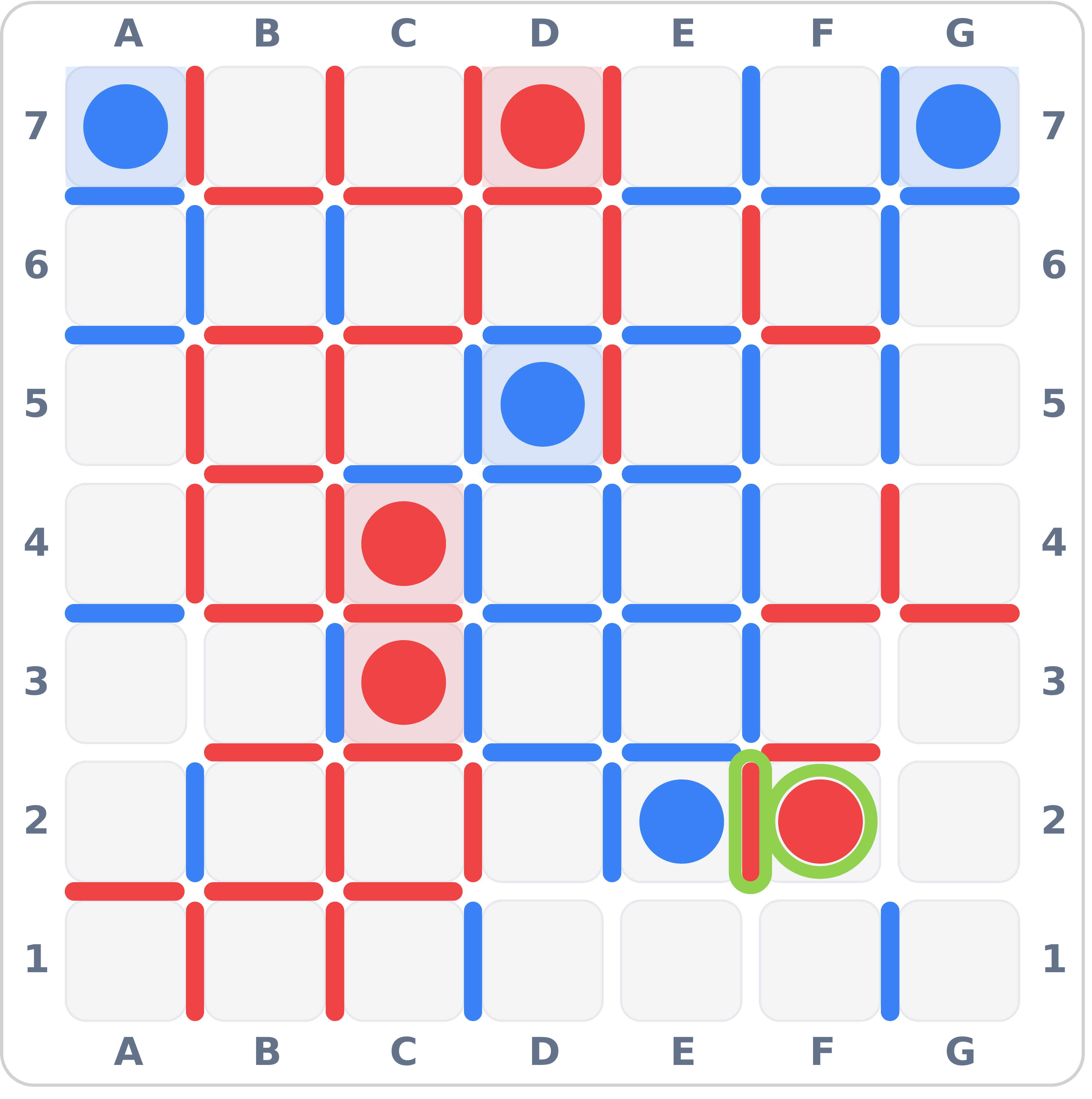}\label{fig:step77}
    }
    \subfloat[Step 78\\(R:8, B:8)]{
        \includegraphics[width=0.18\columnwidth]{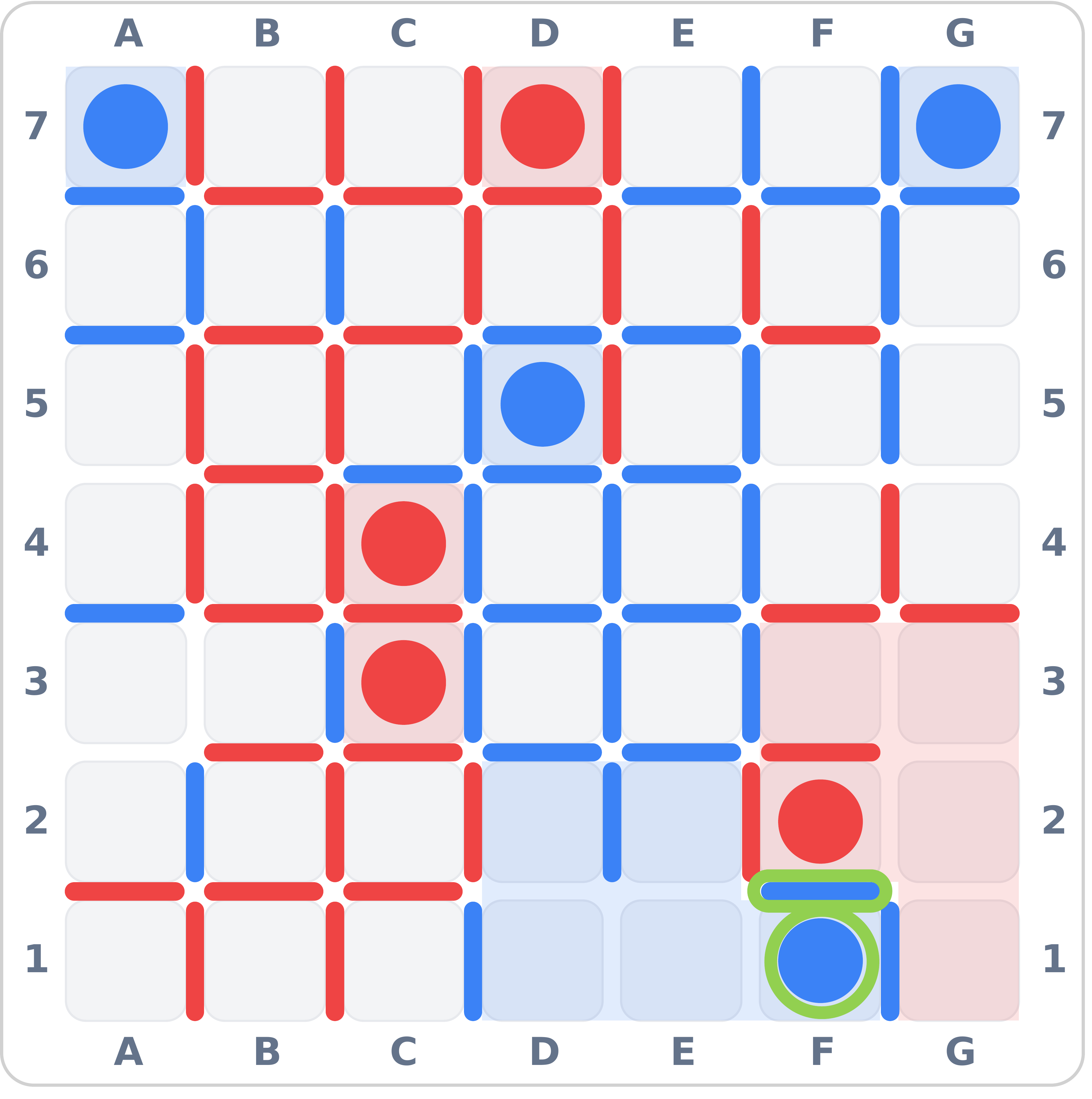}\label{fig:step78}
    }
    
    \caption{Pass Strategy.
    (a) Initial position (Blue to move).
    (b)-(c) Outcomes if Blue or Red plays first in the lower-right region, respectively.
    (d)-(h) Sequence illustrating the game progressing to a draw.
    (R:x, B:y) indicate territory scores.}
    \label{fig:strategy_draw}
\end{figure}

Interestingly, we observe that WallZero develops an implicit passing technique.
Figure~\ref{fig:step53} shows a late-game position where Blue leads by one point, with 10 points in the lower-right region remaining unclaimed.
If Blue plays in that region first (Figure~\ref{fig:step53_blue_final}), it can obtain at most 3 points, leading to a loss.
In contrast, if Red plays first (Figure~\ref{fig:step53_red_final}), both players secure 5 points, allowing Blue to win.
Thus, both players prefer the opponent to play in this region first.

To achieve this, Blue selects the stone at D4, does not move, and builds a wall below (Figure~\ref{fig:step54}), leaving the territory unchanged while effectively transferring the move to Red.
Red responds similarly by sacrificing one point of its own territory, as shown in Figure~\ref{fig:step55}.
Both players continue sacrificing one point each turn to simulate passing until only minimal outer territory remains (Figure~\ref{fig:step76}).
In this case, Red is eventually forced to play in the region first, resulting in a draw (Figure~\ref{fig:step78}).
This example demonstrates an implicit passing technique in WallGo: players may sacrifice small amounts of territory to manipulate turn order, which can be crucial in the endgame.

\section{Discussion}

We present WallZero, an AlphaZero-based WallGo agent with tailored feature design. 
WallZero demonstrated strong performance against professional Go players.
Beyond playing strength, we examined game balance and found that the 4-stone mode improves fairness over the empty mode.
Furthermore, by analyzing the agent's behaviors, we provide deeper insights into strategic principles for mastering WallGo.
Through WallZero, we demonstrate how AlphaZero-based agents can be used not only to achieve strong play but also to quantify balance and extract strategic principles in newly introduced games.

For future work, several extensions are worth exploring. 
One direction is to study the four-player WallGo setting, which requires adaptations of multi-player search methods~\cite{nijssen_enhancements_2011,nijssen_search_2013}.
Another direction is to investigate different board sizes, such as $6\times6$ or $8\times8$, to analyze generalization under varying complexities.
Finally, extending evaluation to a larger pool of human players could provide further insights into learned strategies.

\section*{Acknowledgement}
This research is partially supported by the National Science and Technology Council (NSTC) of the Republic of China (Taiwan) under Grant Number NSTC 113-2221-E-001-009-MY3, NSTC 114-2634-F-A49-004, NSTC 114-2221-E-A49-005, and NSTC 114-2221-E-A49-006. 
The authors would also like to thank the help from two Taiwanese professional Go players, Wei Huang (3-dan) and Chun-Hsun Chou (9-dan).

\bibliographystyle{splncs04}
\bibliography{reference}

@misc{_top_,
  title = {Top 10 {{Non-English Shows}} on {{Netflix Right Now}}},
  journal = {Netflix Tudum},
  howpublished = {https://www.netflix.com/tudum/top10/tv-non-english}
}

@misc{chu_schaoss_2025,
  title = {Schaoss/Wall-Go},
  author = {Chu, Gary},
  year = 2025,
  month = nov,
  copyright = {MIT}
}

@inproceedings{coulom_efficient_2007,
  title = {Efficient {{Selectivity}} and {{Backup Operators}} in {{Monte-Carlo Tree Search}}},
  booktitle = {Computers and {{Games}}},
  author = {Coulom, R{\'e}mi},
  year = 2007,
  series = {Lecture {{Notes}} in {{Computer Science}}},
  pages = {72--83},
  publisher = {Springer},
  address = {Berlin, Heidelberg}
}

@incollection{czech_representation_2024,
  title = {Representation {{Matters}} for {{Mastering Chess}}: {{Improved Feature Representation}} in {{AlphaZero Outperforms Switching}} to {{Transformers}}},
  booktitle = {{{ECAI}} 2024},
  author = {Czech, Johannes and Blüml, Jannis and Kersting, Kristian and Steingrimsson, Hedinn},
  year = 2024,
  pages = {2378--2385},
  publisher = {IOS Press}
}

@inproceedings{kocsis_bandit_2006,
  title = {Bandit {{Based Monte-Carlo Planning}}},
  booktitle = {European {{Conference}} on {{Machine Learning}} and {{Principles}} and {{Practice}} of {{Knowledge Discovery}} in {{Databases}}},
  author = {Kocsis, Levente and Szepesv{\'a}ri, Csaba},
  year = 2006,
  month = sep,
  volume = {2006},
  pages = {282--293}
}

@incollection{nijssen_enhancements_2011,
  title = {Enhancements for {{Multi-Player Monte-Carlo Tree Search}}},
  booktitle = {Computers and {{Games}}},
  author = {Nijssen, J. A. M. and Winands, Mark H. M.},
  year = 2011,
  volume = {6515},
  pages = {238--249},
  publisher = {Springer Berlin Heidelberg},
  address = {Berlin, Heidelberg}
}

@article{nijssen_search_2013,
  title = {Search {{Policies}} in {{Multi-Player Games}}},
  author = {Nijssen, J. A. M. and Winands, Mark H. M.},
  year = 2013,
  month = mar,
  journal = {ICGA Journal},
  volume = {36},
  number = {1},
  pages = {3--21}
}

@article{silver_general_2018,
  title = {A General Reinforcement Learning Algorithm That Masters Chess, Shogi, and {{Go}} through Self-Play},
  author = {Silver, David and Hubert, Thomas and Schrittwieser, Julian and Antonoglou, Ioannis and Lai, Matthew and Guez, Arthur and Lanctot, Marc and Sifre, Laurent and Kumaran, Dharshan and Graepel, Thore and Lillicrap, Timothy and Simonyan, Karen and Hassabis, Demis},
  year = 2018,
  month = dec,
  journal = {Science},
  volume = {362},
  number = {6419},
  pages = {1140--1144},
  publisher = {American Association for the Advancement of Science}
}

@article{silver_mastering_2017,
  title = {Mastering the Game of {{Go}} without Human Knowledge},
  author = {Silver, David and Schrittwieser, Julian and Simonyan, Karen and Antonoglou, Ioannis and Huang, Aja and Guez, Arthur and Hubert, Thomas and Baker, Lucas and Lai, Matthew and Bolton, Adrian and Chen, Yutian and Lillicrap, Timothy and Hui, Fan and Sifre, Laurent and {van den Driessche}, George and Graepel, Thore and Hassabis, Demis},
  year = 2017,
  month = oct,
  journal = {Nature},
  volume = {550},
  number = {7676},
  pages = {354--359},
  publisher = {Nature Publishing Group},
  copyright = {2017 Macmillan Publishers Limited, part of Springer Nature. All rights reserved.}
}

@misc{team_play_,
  title = {Play {{Wall Go Online}} \textbar{} {{Strategic Board Game}} vs {{AI}} \& {{Multiplayer}}},
  author = {Team, Wall Go},
  howpublished = {https://playwallgo.com}
}

@misc{tomasev_assessing_2020,
  title = {Assessing {{Game Balance}} with {{AlphaZero}}: {{Exploring Alternative Rule Sets}} in {{Chess}}},
  author = {Toma{\v s}ev, Nenad and Paquet, Ulrich and Hassabis, Demis and Kramnik, Vladimir},
  year = 2020,
  month = sep
}

@article{wang_evaluating_2025,
  title = {{Evaluating Game Difficulty in Tetris Block Puzzle}},
  author = {Wang, Chun-Jui and Guo, Jian-Ting and Guei, Hung and Shih, Chung-Chih and Wu, Ti-Rong and Wu, I-Chen},
  year = 2025,
  month = nov,
  journal = {The 30th Game Programming Workshop (GPW-25)},
  volume = {2025},
  pages = {54--59},
}

@inproceedings{wu_accelerating_2020,
  title = {Accelerating {{Self-Play Learning}} in {{Go}}},
  booktitle = {Proceedings of the {{AAAI Workshop}} on {{Reinforcement Learning}} in {{Games}}},
  author = {Wu, David J.},
  year = 2020,
  month = nov
}

@article{wu_minizero_2025,
  title = {{{MiniZero}}: {{Comparative Analysis}} of {{AlphaZero}} and {{MuZero}} on {{Go}}, {{Othello}}, and {{Atari Games}}},
  author = {Wu, Ti-Rong and Guei, Hung and Peng, Pei-Chiun and Huang, Po-Wei and Wei, Ting Han and Shih, Chung-Chin and Tsai, Yun-Jui},
  year = 2025,
  month = mar,
  journal = {IEEE Transactions on Games},
  volume = {17},
  number = {1},
  pages = {125--137}
}

\end{document}